\documentclass[journal]{IEEEtran}
\usepackage{amsmath,amsfonts}
\usepackage{cite}
\usepackage{amsmath,amssymb,amsfonts}
\usepackage[linesnumbered,ruled,vlined]{algorithm2e}
\usepackage{array}
\usepackage[caption=false,font=normalsize,labelfont=sf,textfont=sf]{subfig}
\usepackage{textcomp}
\usepackage{stfloats}
\usepackage{url}
\usepackage{verbatim}
\usepackage{graphicx}
\usepackage{hyperref}
\usepackage{booktabs}
\usepackage{multirow}
\usepackage{url}
\usepackage{amsmath}
\usepackage{tikz}
\newcommand*\circled[1]{\tikz[baseline=(char.base)]{
            \node[shape=circle,draw,inner sep=2pt] (char) {#1};}}

\begin{document}

\title{Weakly Augmented Variational Autoencoder in Time Series Anomaly Detection}


\author{Zhangkai Wu, Longbing Cao,~\IEEEmembership{Senior Member, ~IEEE, Qi Zhang, Junxian Zhou, Hui Chen }
\thanks{The work is partially sponsored by Australian Research Council Discovery and Future Fellowship grants (DP190101079 and FT190100734).}
\thanks{ Zhangkai Wu, Junxian Zhou with the School of Computer Science, the University of Technology Sydney, 15 Broadway, Ultimo 2007, NSW, Australia. (E-mail: berenwu1938@gmail.com, junxian.zhou@student.uts.edu.au)}
\thanks{Hui Chen and Longbing Cao are with the DataX Research Centre and School of Computing, Macquarie University, NSW 2109, Australia. (email: hui.chen2@students.mq.edu.au, longbing.cao@mq.edu.au)}
\thanks{Qi Zhang is with the Department of Computer Science, Tongji University, Shanghai 201804, China. (E-mail: zhangqi\_cs@tongji.edu.cn)}

}



\maketitle

\begin{abstract}
Due to their unsupervised training and uncertainty estimation, deep Variational Autoencoders (VAEs) have become powerful tools for reconstruction-based Time Series Anomaly Detection (TSAD). Existing VAE-based TSAD methods, either statistical or deep, tune \textit{meta-priors} to estimate the likelihood probability for effectively capturing spatiotemporal dependencies in the data. However, these methods confront the challenge of inherent data scarcity, which is often the case in anomaly detection tasks. Such scarcity easily leads to \textit{latent holes}, discontinuous regions in latent space, resulting in non-robust reconstructions on these discontinuous spaces. We propose a novel generative framework that combines VAEs with self-supervised learning (SSL) to address this issue. Our framework augments latent representation to mitigate the disruptions caused by anomalies in the low-dimensional space, aiming to lead to corresponding more robust reconstructions for detection.

This framework marks a significant advancement in VAE design by integrating SSL to refine likelihood enhancement. Our proposed VAE model, specifically tailored for TSAD, augments latent representations via enhanced training, increasing robustness to normal data likelihoods and improving sensitivity to anomalies. Additionally, we present a practical implementation of this conceptual framework called the Weakly Augmented Variational Autoencoder (WAVAE), which directly augments input data to enrich the latent representation. This approach synchronizes the training of both augmented and raw models and aligns their convergence in data likelihood optimization space. To achieve this, we maximize mutual information within the Evidence Lower Bound (ELBO), utilizing contrastive learning for shallow learning and a discriminator-based adversarial strategy for deep learning.

Extensive empirical experiments on five public synthetic and real datasets validate the efficacy of our framework. These experiments provide compelling evidence of the superior performance of our approach in TSAD, as demonstrated by achieving higher ROC-AUC and PR-AUC scores compared to state-of-the-art models. Furthermore, we delve into the nuances of VAE model design and time series preprocessing, offering comprehensive ablation studies to examine the sensitivity of various modules and hyperparameters in deep optimization.

\end{abstract}

\begin{IEEEkeywords}
Variational Autoencoder, Time Series Anomaly Detection, Self Supervised Learning, Data Augmentation, Contrast Learning, Adversarial Learning.
\end{IEEEkeywords}

\section{Introduction}
\label{sec:intro}

\IEEEPARstart{D}{eep} probabilistic generative models have revolutionized unsupervised data generation by leveraging the neural network's universal approximation theorem. This innovation manifests in various forms, such as encoding-decoding mechanisms \cite{kingma2013auto}, diffusion-denoise processes \cite{ho2020denoising}, and sender-receiver in compression \cite{graves2023bayesian}. Notably, VAEs have emerged as a central focus in this domain. Deep VAEs empowered by large-scale neural networks \cite{vahdat2020nvae} and the organization of semantic representations \cite{takida2022sq, manduchi2023tree} have demonstrated exceptional capabilities in reconstructing and generating multimodal data, including images \cite{razavi2019generating,wu2023evae}, tabular \cite{xu2019modeling, kotelnikov2023tabddpm}, and time series \cite{zhu2023markovian, jin2022pfvae}. Owing to their ability to learn generative factors in continuous and smooth low-dimensional space while fitting data likelihoods, VAEs exhibit robust representation learning capabilities in disentanglement \cite{liu2023c,wu2023c}, classification \cite{xiao2023trading}, and clustering \cite{tonekaboni2022decoupling}. In particular, VAEs are designed to estimate likelihood distribution learned from most normal samples as a detector, providing an unsupervised and interpretable paradigm for anomaly detection. The underlying assumption is that unknown anomaly patterns typically exhibit statistical characteristics that deviate significantly from the normal distribution. 

\begin{figure}[htbp]
    \centering
    \includegraphics[width=1\columnwidth]{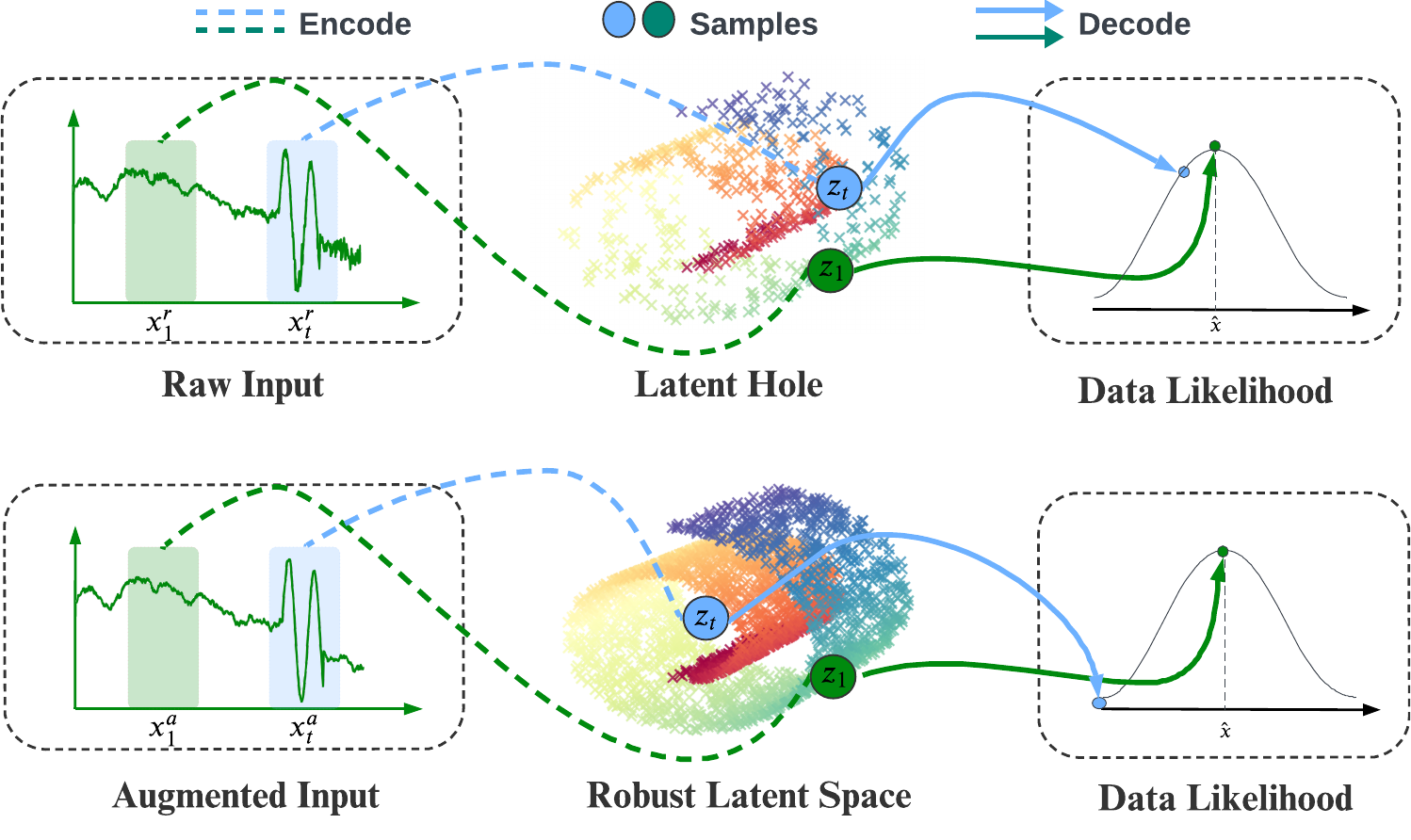}
    \caption{ Comparison of the \textit{latent hole} phenomenon induced by anomalies in Nonrobust VAE-based TSAD Models (upper section) with the robust representation learning space fostered by WAVAE (lower section). The upper part of the figure delineates the rise of the \textit{latent hole} within the Nonrobust TSAD model and its effect on model robustness. Specifically, anomalous sequences $\boldsymbol{x}_{t}^{r}$ (depicted within the blue window in the upper section), when encoded into the representation space, disrupt the structural integrity of the latent space. This disruption results in \textit{latent hole} primarily because these anomalous sequences $\boldsymbol{x}_{t}^{r}$ lack the spatiotemporal coherence inherent in the normal sequence $\boldsymbol{x}_{1}^{r}$. Consequently, sampling from these discontinuous regions leads to a mismatch between the representation (indicated by the blue dot $\boldsymbol{z}_{t}$) and its generation (also shown by the blue dot in the likelihood function), as illustrated in the upper section, a disproportionately high likelihood function mass characterizes the representation in the latent space. In such scenarios, the TSAD model may erroneously classify an anomaly as normal, compromising its robustness. In contrast, the lower section demonstrates how data augmentation via the WAVAE model can engender a more continuous and smoothly distributed data likelihood (as depicted in the central part of the bottom figure). In this context, representation $\boldsymbol{z}_{t}$) encoded by anomalous sequences $\boldsymbol{x}_{t}^{a}$ sampled from regions outside the normal latent space are associated with a lower likelihood function mass, thereby enhancing the robustness and efficacy of anomaly detection in TSAD tasks.}
    \label{fig:latent}
\end{figure}

Recent research has shown a growing preference for VAEs in Time Series Anomaly Detection (TSAD), particularly those that integrate \textit{meta-priors}\cite{tschannen2018recent} into their design.  These methods have been validated to be effective and crucial in capturing the spatiotemporal dependencies within data, thereby enhancing data likelihood modeling. Specifically, these models often assume certain \textit{meta-priors}, e.g., latent structures, which are crafted using deep learning or probabilistic tools. The goal is to accurately represent the likelihood of most data points, enabling the models to detect anomalies effectively. For instance,  time-varying priors that adapt to dynamic assumptions~\cite{chung2015recurrent, park2018multimodal, su2019robust, kieu2022anomaly} have demonstrated effectiveness and powerful capabilities in capturing sequence data likelihoods. Additionally, other studies \cite{lai2023nominality,DBLP:conf/icml/LiCCWTZ23} have proposed the creation of task-specific priors based on a factorized assumption explicitly designed to model contextual dependence structure in latent space. Various strategies have been employed to achieve this, such as using prototype distribution-based representations optimized through meta-learning and decomposing contextual representations.
Whether meta-priors are integrated implicitly or explicitly, these methods are fundamentally rooted in model-based designs. However, they often cater to specific scenarios and face training challenges stemming from the data scarcity issue in deep learning or statistical estimation techniques. Additionally, these methods overlook the effective utilization of data, which becomes particularly problematic when modeling small sequence datasets in real-world time-series TSAD scenarios. The issue in VAE-based models easily leads to \textit{latent holes} \cite{khan2023adversarial}, i.e., discontinuous regions in latent space, resulting in non-robust reconstruction \cite{zhou2017anomaly,akrami2019robust,eduardo2020robust,cao2022coupled,kieu2022anomaly}. The \textit{latent holes} issue occurs when the encoder in these models maps unknown anomalies into the latent space without being adequately corrected by enough normal data. Since these anomalies lack the spatiotemporal properties of normal time series data, they disrupt the formation of a continuous and smooth latent space for normal samples. Consequently, representations sampled from these \textit{latent holes} fail to accurately reconstruct the input samples, causing a discrepancy between the representations and the reconstructed data. This mismatch significantly impairs the anomaly detector's performance and compromises the model's overall robustness. For a more intuitive understanding of this phenomenon, refer to Fig. \ref{fig:latent}.

In light of these challenges, we propose to improve data utilization using self-supervised learning (SSL) to enhance representation learning and induce latent space robustness. SSL \cite{zhang2023self} enables models to extract more informative representations from unlabeled data, leading to sufficient training. To achieve this, we employ data augmentation on unlabeled datasets through SSL strategies, facilitating the training of models through contrastive or adversarial methods for the TSAD task.
Our contributions can be summarized as:
\begin{itemize}
    \item \textbf{Generative self-supervised learning framework for TSAD}: We present an enhanced generative framework using self-supervised learning. We define a likelihood function for learning and the derivation of a surrogate error for optimization. This novel approach sets the stage for more effective model design in VAE-based TSAD.
    \item \textbf{Deep and shallow learning in augmented models}: Building upon this framework, we implement the Weakly Augmented Variational Autoencoder (WAVA) that incorporates data augmentation, enabling the model to undergo thorough training with support from augmented counterparts. We have also devised two distinct learning approaches, deep and shallow, to integrate these two models effectively.
    \item \textbf{State-of-the-art performance}: Extensive experiments on five public datasets demonstrate the effectiveness of our approach. We achieved superior performance in ROC-AUC and PR-AUC, surpassing state-of-the-art models. Additionally, we provide comprehensive ablation studies delving into the design of the VAE model, time series preprocessing, and sensitivity analysis on different modules and hyperparameters in deep optimization.    
\end{itemize}

\begin{figure}[htbp]
    \centering
    \includegraphics[width=1\columnwidth]{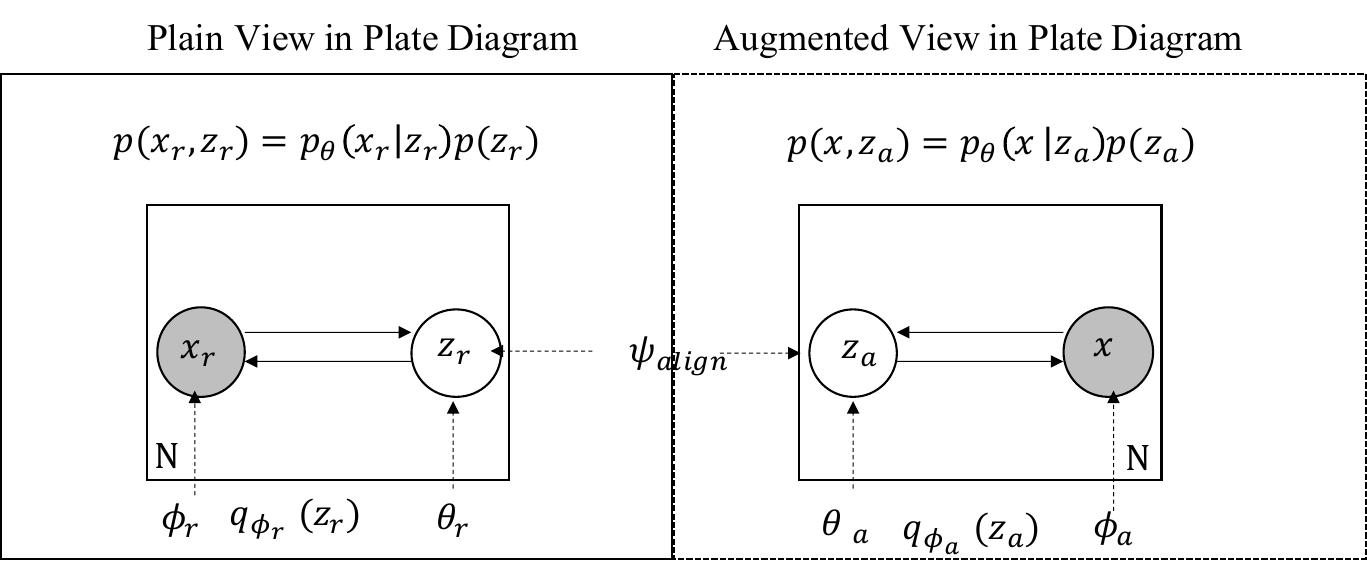}
    \caption{Graphical Model for Augmented Variational Autoencoders. Under the plate notation rules, a white circle denotes a hidden (or latent) variable, while a gray circle signifies an observed variable. The variables contained within the square denote local variables, which are independently repeated $N$ times. Dashed arrow edges imply conditional dependence. Dotted lines represent parameters. Referring to the plate diagram, it is evident that our methodology encompasses the utilization of two generative models. The inference part of models, i.e., $q_{\phi_{\mathrm{r}}}$ and $q_{\phi_{\mathrm{a}}}$, encode the raw input, denoted as $\boldsymbol{x}_{\mathrm{r}}$, and the augmented input, $\boldsymbol{x}_{\mathrm{a}}$, into their respective low-dimensional representations, $\boldsymbol{z}_{\mathrm{r}}$ and $\boldsymbol{z}_{\mathrm{a}}$. Subsequently, the generative parts of models $p_{\theta_{\mathrm{r}}}$ and $p_{\theta_{\mathrm{a}}}$, sample the latent space reconstruct the input samples, respectively. We employ a $\psi$ parameterized module to synchronize the learning outcomes of both models.}
    \label{fig:plate}
\end{figure}

\section{Backgroud}
\label{sec:bg}

\subsection{Generative model-based Time Series Anomaly Detection}
\subsubsection{Implicit Data Fitting by Non-probabilistic Generative Models} Non-probabilistic generative model-based algorithms for TSAD aim to reconstruct data robustly. Prior works have concentrated on optimizing this reconstructing process to match the characteristics of time-series data via the design of deep network embeddings. Specifically, \cite{chen2017outlier} and \cite{kieu2018outlier} implemented an Autoencoder (AE) framework, deploying symmetric encoder-decoder structures and assembling one or multiple CNN-based encoder-decoders for the reconstruction of sequence data. Furthermore, \cite{zhou2019beatgan} utilized a Generative Adversarial Network (GAN)-based reconstruction for anomaly detection, implicitly fitting a likelihood function based on the normal data through an adversarial mechanism.

\subsubsection{Explicit Data Fitting by Probabilistic generative Models} Unlike AE-based models that learn an encoding-decoding process for datasets, VAE-based models excel in identifying continuous representations within a low-dimensional space. These representations, characterized by their smooth and continuous nature in the hidden space, are essential for preserving probabilistic properties during sampling. Consequently, VAEs can reconstruct samples with increased sharpness and interpretability, outperforming their autoencoder-based counterparts. In contrast to GANs, VAEs explicitly model the data's likelihood distribution and provide additional constraints on the data's posterior distribution based on a preset prior, making them more suitable for modeling data in dynamic areas and designing end-to-end anomaly detection tasks. For instance, \cite{liao2018unified} utilizes a Gaussian Mixture Model (GMM) assumption for data likelihood distribution, and \cite{xu2018unsupervised} employs a dynamical prior over time. 

 \textbf{Issues in VAE based TSAD:} VAEs tend to sacrifice representation \cite{zhao2017infovae} for data fitting. In that case, the induced latent hole will lead to the lack of robustness Represented by \textit{latent hole}. At the same time, the model's failure to learn the likelihood of the sequence data exacerbates its robustness issues. Specifically, VAE-based anomaly detection algorithms typically employ a Convolutional Neural Network (CNN) architecture for data encoding. While effective for image data, this approach often fails to capture the temporal characteristics of time-series data, such as seasonality, periodicity, and frequency domain features, through CNN encoding filters. The shallow Fully Connected Network (FCN) networks are employed in VAEs as substitutes. As a result, the naive structure cannot capture varying dependence, and compared to the image data, the sequence in training is relatively small. Due to the model and data issue, the generative modal cannot converge to the optimal.

 
 \textbf{Advances in VAE-based TSAD:} To remedy this, the traditional variational framework has been revoluted, integrating the \textit{meta-prior} in generative modeling. For instance, a Variational Recurrent Neural Network (VRNN) has been proposed, establishing a model for the variational autoencoder's inference, prior, and reconstruction processes by capturing the temporal dependencies between intermediate variable $\boldsymbol{h}$ in the deterministic model and input variables $\boldsymbol{x}$ in the recurrent neural network. This approach and its variants \cite{chung2015recurrent,su2019robust,kieu2022anomaly} effectively utilize the variational autoencoder to learn and model the latent distribution of data while maintaining the temporal dependence of the recurrent neural network. On the other hand, the variational representation can be designed. \cite{DBLP:conf/icml/LiCCWTZ23} utilizes prototype-based approaches to define latent representations for Multivariate Time Series (MTS) and learn a robust likelihood distribution of normal data.


\subsection{Self-supervised Learning on Time Series Data in Determistic and Generative Models}
In deterministic models, augmenting time-series data or their representations, combined with specific self-supervised algorithms, can provide sufficient depth for training in downstream tasks. For instance, in prediction tasks, \cite{woo2022cost} encodes time-series segments in both time and frequency domains to obtain positive and negative sample pairs, using contrastive learning to capture the seasonal-trend representation of time-series data. \cite{hou2022multi} constructs positive pairs with multi-granularity time-series segments and corresponding latent variable representations, enhancing fine-grained information for prediction by maximizing mutual information.
\par
In classification tasks, \cite{lee2022self} forms pairwise representations of global and local input series, obtaining informational gains through adversarial learning. For anomaly detection tasks, \cite{xu2021anomaly} aims to acquire spatio-temporal dependent representations suitable for downstream tasks. \cite{yue2022ts2vec} proposes a multi-layer representation learning framework to obtain consistent, contextual representations of overlapping segments, designing a contrastive loss by decomposing overlapping subsequences in both instance and temporal dimensions to obtain positive and negative sample pairs. Additionally, \cite{yang2022unsupervised} employs a dual bilinear process at the encoding level to capture positive and negative samples of time sequences, thereby capturing both long and short-term dependencies.
\par
In contrast, SSL based on time-series generative models typically focuses on data and representation augmentation as a generative approach. For instance, Autoencoder (AE) based methods, such as those presented in \cite{wang2022learning,chen2022deep,shukla2021heteroscedastic,zhang2022grelen}, leverage the AE architecture for data augmentation. Similarly, diffusion-based approaches, as seen in \cite{li2022generative,alcaraz2022diffusion,wen2023diffstg}, employ diffusion processes to augment time-series data and representations.

\section{Augumentation Guided Generative Anomaly Detection}
\label{sec:method}

In this section, we first provide the problem definition for generative model-based TSAD. Subsequently, we depict the structure of the augmented generative model in the form of a plate diagram, illustrating the random variables and their dependency structure. The augmented guided generative anomaly detection model operates within the framework of probabilistic generative models, employing self-supervised techniques to augment the latent variables $\boldsymbol{z}$ during the training process of the generative model, thereby enhancing the deep model's fit to the data likelihood. Herein, we have implemented a self-supervised variational autoencoder based on input data augmentation, which preprocesses the input data $\boldsymbol{x}$ to generate latent variables $\boldsymbol{z}_{\mathrm{a}}$ with different views. To align the likelihood functions of the raw and augmented models, we have developed two distinct mutual information loss functions, one grounded in depth and the other in statistics. By maximizing the mutual information between them, we draw the models closer to fitting the same distribution.

\subsection{Problem Definition}

Time series data is succinctly represented as $\mathcal{X} := \{(\boldsymbol{x}^{(i)}, \boldsymbol{y}^{(i)})\}_{i=1}^n$, encompassing $n$ time-stamped observations $\boldsymbol{x} \in \mathbb{R}^c$ situated within a $c$-dimensional representation space, each paired with a discrete observation $\boldsymbol{y}$. The observation $\boldsymbol{y}$ is assigned discrete values across $l$ predefined classes, delineated as $\boldsymbol{y} \in \{0, 1, \ldots, l-1\}$. Here, $c$ denotes the feature dimensionality at each time point, categorizing the dataset as a Multivariate Time Series (MTS) when $c > 1$ and as a Univariate Time Series (TS) for $c = 1$. In all figures and equations, to enhance notational conciseness, we propose abbreviating $\mathrm{raw}$ as $\mathrm{r}$ and $\mathrm{augmentation}$ as $\mathrm{a}$.

In generative model-based TSAD, the focus is on learning a reconstructing model, i.e., $\mathcal{M}_{\mathrm{normal}}$, that models the mass of loglikelihood of the majority of normal data points within the entire dataset $\mathcal{X} = \{\mathcal{X}_{\mathrm{normal}},\mathcal{X}_{\mathrm{abnormal}}\}$. Anomalies are then identified in an unsupervised, end-to-end fashion by calculating the anomaly score, denoted as $AS(\boldsymbol{x},\hat{\boldsymbol{x}})$, which quantifies the difference between a given input $\boldsymbol{x}$ and its modeled copy $\hat{\boldsymbol{x}}$ as reconstructed by $\mathcal{M}_{\mathrm{normal}}$. This approach is feasible, assuming that the log-likelihood learned from normal observations will diverge notably when encountering anomalous data, yielding elevated anomaly scores.

\subsection{Data Augmentation Guided Probabilistic Generative Model}

The plate diagram in Fig. \ref{fig:plate} defines an augmented-based probabilistic generative model. The upper part of the diagram specifies the learned joint distribution of the original data $\boldsymbol{x}_{\mathrm{r}}$ and its latent variable $\boldsymbol{z}_{\mathrm{r}}$, i.e., $p(\boldsymbol{x}_{\mathrm{r}},\boldsymbol{z}_{\mathrm{r}})$, where the latent variable $\boldsymbol{z}_{\mathrm{r}}$ is generated by an inference network $q_{\phi_{\mathrm{r}}}(\boldsymbol{z}_{\mathrm{r}}|\boldsymbol{x}_{\mathrm{r}})$ parameterized by $\phi_{\mathrm{r}}$, and the reconstructed variable $\hat{x}_{\mathrm{r}}$ is produced by a generative network $p_{\theta_{\mathrm{r}}}(\boldsymbol{x}_{\mathrm{r}}|\boldsymbol{z}_{\mathrm{r}})$ parameterized by $\theta_{\mathrm{r}}$. The model optimizes an approximate surrogate error $\mathcal{L}^{\mathrm{r}}_{\mathrm{ELBO}}$, comprising a reconstruction loss that maximizes the likelihood distribution $\mathcal{L}_{\mathrm{R}}^{\mathrm{r}}$ and a $D_{\mathrm{KL}}$ loss that minimizes the discrepancy between the prior of the latent variables and their variational posterior $\mathcal{L}_{\mathrm{I}}^{\mathrm{r}}$, i.e.
\begin{equation}
\begin{aligned}
\mathcal{L}^{\mathrm{r}}_{\mathrm{ELBO}} := & \underbrace{\mathbb{E}_{q_{\phi_{\mathrm{r}}}(\boldsymbol{z}_{\mathrm{r}} \mid \boldsymbol{x}_{\mathrm{r}})}\Big[\log p_{\theta_{\mathrm{r}}}(\boldsymbol{x}_{\mathrm{r}} \mid \boldsymbol{z}_{\mathrm{r}})}_{\mathcal{L}_{\mathrm{R}}^{\mathrm{r}}}\Big] \\
 & - \beta \underbrace{D_{K L}\big(q_\phi(\boldsymbol{z}_{\mathrm{r}} \mid \boldsymbol{x}_{\mathrm{r}}) || p(\boldsymbol{z})}_{\mathcal{L}_{\mathrm{I}}^{\mathrm{r}}}\big).
\end{aligned}
\label{eq:raw_eblo}
\end{equation}
The lower part of the plate diagram outlines the probabilistic model of the joint distribution of the augmented view data $\boldsymbol{x}_{\mathrm{a}}$ and latent variables $\boldsymbol{z}_{\mathrm{a}}$, i.e., $p(\boldsymbol{x}_{\mathrm{a}},\boldsymbol{z}_{\mathrm{a}})$ with latent variables $\boldsymbol{z}_{\mathrm{a}}$ derived from an augmented inference network $q_{\phi_{\mathrm{a}}}(\boldsymbol{z}_{\mathrm{a}}|\boldsymbol{x})$, i.e., $\boldsymbol{z}_{\mathrm{a}} \sim q_{\phi_{\mathrm{a}}}(\boldsymbol{z}_{\mathrm{a}}|\boldsymbol{x})$. Similar to the above, the model optimizes an augmented reconstruction loss $\mathcal{L}_{\mathrm{R}}^{\mathrm{a}}$ and inference loss $\mathcal{L}_{\mathrm{I}}^{\mathrm{a}}$, i.e.,
\begin{equation}
\begin{aligned}
\mathcal{L}^{\mathrm{a}}_{\mathrm{ELBO}} := & \underbrace{\mathbb{E}_{q_{\phi_{\mathrm{a}}}(\boldsymbol{z}_{\mathrm{a}} \mid \boldsymbol{x}_{\mathrm{a}})}\Big[\log p_{\theta_{\mathrm{a}}}(\boldsymbol{x}_{\mathrm{a}} \mid \boldsymbol{z}_{\mathrm{a}})}_{\mathcal{L}_{\mathrm{R}}^{\mathrm{a}}}\Big] \\
 & - \beta \underbrace{D_{\mathrm{KL}}\big(q_{\phi}(\boldsymbol{z}_{\mathrm{a}} \mid \boldsymbol{x}_{\mathrm{a}}) || p(\boldsymbol{z})}_{\mathcal{L}_{\mathrm{I}}^{\mathrm{a}}}\big).
\end{aligned}
\label{eq:aug_eblo}
\end{equation}
On the one hand, both models strive to fit their respective data distribution likelihoods. On the other, we leverage the advantage of data augmentation by maximizing the mutual information $I(\boldsymbol{z}_{\mathrm{r}},\boldsymbol{z}_{\mathrm{a}})$ between two latent models, optimizing a mutual information loss parameterized by $\psi$ (in deep learning approximation), to train the models for maximal data likelihood synergistically.

Given the variety of latent variable augmentations, this paper augments the raw data to augment the model. In summary, we propose an augmented probabilistic generative model to learn a joint likelihood function $p(\boldsymbol{x}_{\mathrm{r}}, \boldsymbol{z}_{\mathrm{r}}, \boldsymbol{x}_{\mathrm{a}},\boldsymbol{z}_{\mathrm{a}})$ for anomaly detection while simultaneously optimizing an inference network parameterized by $\phi_{\mathrm{r}},\phi_{\mathrm{a}}$, a generative network parameterized by $\theta_{\mathrm{r}},\theta_{\mathrm{a}}$, and an alignment network parameterized by $\psi$. The $\psi$ can be parameterized by neurons in deep learning approximation and pseudo-parameters in shallow learning. The generative process is as follows:

\begin{equation}
 p\left(\boldsymbol{x}_{\mathrm{r}}, \boldsymbol{x}_{\mathrm{a}}\right)= \int p\left(\boldsymbol{x}_{\mathrm{r}}, \boldsymbol{x}_{\mathrm{a}}, \boldsymbol{z}_{\mathrm{r}}, \boldsymbol{z}_{\mathrm{a}}\right) \mathrm{d} \boldsymbol{z}_{\mathrm{r}} \mathrm{d} \boldsymbol{z}_{\mathrm{a}},
\label{eq:joint}
\end{equation}
where $\boldsymbol{x}_{\mathrm{r}}$ represents the raw input datapoint, $\boldsymbol{x}_{\mathrm{a}}$ is the augmented sample based on the input, $\boldsymbol{z}_{\mathrm{r}}$ is the raw latent variable, and $\boldsymbol{z}_{\mathrm{a}}$ is the augmented latent variable.

The joint distribution is often too high-dimensional and sophisticated to solve it directly. To address this, a tractable variational distribution $q(\boldsymbol{z}_{\mathrm{r}},\boldsymbol{z}_{\mathrm{a}})$ is employed as an approximation within the framework of Variational Inference (VI). Due to the computational convenience it offers, we typically take the logarithm of the distribution. Consequently, as depicted in Equation \ref{eq:joint}, the likelihood of data that encompasses latent variables can be decomposed as follows:
\begin{equation}
\begin{aligned}
 & p\left(\boldsymbol{x}_{\mathrm{r}}, \boldsymbol{x}_{\mathrm{a}}\right) \\  =& \int \frac{p\left(\boldsymbol{x}_{\mathrm{r}}, \boldsymbol{x}_{\mathrm{a}}, \boldsymbol{z}_{\mathrm{r}}, \boldsymbol{z}_{\mathrm{a}}\right) q(\boldsymbol{z}_{\mathrm{r}},\boldsymbol{z}_{\mathrm{a}})}{q(\boldsymbol{z}_{\mathrm{r}},\boldsymbol{z}_{\mathrm{a}})} \mathrm{d} \boldsymbol{z}_{\mathrm{r}} \mathrm{d} \boldsymbol{z}_{\mathrm{a}}, \\
\end{aligned}
\label{eq:joint-inter}
\end{equation}
\par
and we can get the $\log$ versions as follows:
\begin{equation}
\begin{aligned}
& \log p\left(\boldsymbol{x}_{\mathrm{r}}, \boldsymbol{x}_{\mathrm{a}}\right) \\  =&\log \int \frac{p\left(\boldsymbol{x}_{\mathrm{r}}, \boldsymbol{x}_{\mathrm{a}}, \boldsymbol{z}_{\mathrm{r}}, \boldsymbol{z}_{\mathrm{a}}\right) q(\boldsymbol{z}_{\mathrm{r}},\boldsymbol{z}_{\mathrm{a}})}{q(\boldsymbol{z}_{\mathrm{r}},\boldsymbol{z}_{\mathrm{a}})} \mathrm{d} \boldsymbol{z}_{\mathrm{r}} \mathrm{d} \boldsymbol{z}_{\mathrm{a}}.
\end{aligned}
\label{eq:joint-log}
\end{equation}
Given the $\log$ is a convex function, we can get a lower bound by Jensen's inequality:
\begin{equation}
\begin{aligned}
&  \log p\left(\boldsymbol{x}_{\mathrm{r}}, \boldsymbol{x}_{\mathrm{a}}\right)\\   
=&\log \int \frac{p\left(\boldsymbol{x}_{\mathrm{r}}, \boldsymbol{x}_{\mathrm{a}}, \boldsymbol{z}_{\mathrm{r}}, \boldsymbol{z}_{\mathrm{a}}\right) q(\boldsymbol{z}_{\mathrm{r}},\boldsymbol{z}_{\mathrm{a}})}{q(\boldsymbol{z}_{\mathrm{r}},\boldsymbol{z}_{\mathrm{a}})} \mathrm{d} \boldsymbol{z}_{\mathrm{r}} \mathrm{d} \boldsymbol{z}_{\mathrm{a}} \\
=& \log \mathbb{E}_{q\left(\boldsymbol{z}_{\mathrm{r}}, \boldsymbol{z}_{\mathrm{a}} \mid \boldsymbol{x}_{\mathrm{r}}, \boldsymbol{x}_{\mathrm{a}}\right)}\left[\frac{p\left(\boldsymbol{x}_{\mathrm{r}}, x_{\mathrm{a}}, \boldsymbol{z}_{\mathrm{r}}, \boldsymbol{z}_{\mathrm{a}}\right)}{q\left(\boldsymbol{z}_{\mathrm{r}}, \boldsymbol{z}_{\mathrm{a}} \mid \boldsymbol{x}_{\mathrm{r}}, \boldsymbol{x}_{\mathrm{a}}\right)}\right] \\
\geq &\mathbb{E}_{q\left(\boldsymbol{z}_{\mathrm{r}}, \boldsymbol{z}_{\mathrm{a}} \mid \boldsymbol{x}_{\mathrm{r}}, \boldsymbol{x}_{\mathrm{a}}\right)}\log \left[\frac{p\left(\boldsymbol{x}_{\mathrm{r}}, \boldsymbol{x}_{\mathrm{a}}, \boldsymbol{z}_{\mathrm{r}}, \boldsymbol{z}_{\mathrm{a}}\right)}{q\left(\boldsymbol{z}_{\mathrm{r}}, \boldsymbol{z}_{\mathrm{a}} \mid \boldsymbol{x}_{\mathrm{r}}, \boldsymbol{x}_{\mathrm{a}}\right)}\right] \\
=&\mathbb{E}_{q\left(\boldsymbol{z}_{\mathrm{r}}, \boldsymbol{z}_{\mathrm{a}} \mid \boldsymbol{x}_{\mathrm{r}}, \boldsymbol{x}_{\mathrm{a}}\right)}\log \left[\frac{p(\boldsymbol{x}_{\mathrm{r}} \mid \boldsymbol{z}_{\mathrm{r}}) p\left(\boldsymbol{x}_{\mathrm{a}} \mid \boldsymbol{z}_{\mathrm{a}}\right) p\left(\boldsymbol{z}_{\mathrm{r}}, \boldsymbol{z}_{\mathrm{a}}\right)}{q(\boldsymbol{z}_{\mathrm{r}} \mid \boldsymbol{x}_{\mathrm{r}}) q\left(\boldsymbol{z}_{\mathrm{a}} \mid x_{\mathrm{a}}\right)}\right] \\
=&\underbrace{\mathbb{E}_{q(\boldsymbol{z}_{\mathrm{r}} \mid \boldsymbol{x}_{\mathrm{r}})}\log [p(\boldsymbol{x}_{\mathrm{r}} \mid \boldsymbol{z}_{\mathrm{r}})]+\mathbb{E}_{q\left(\boldsymbol{z}_{\mathrm{a}} \mid \boldsymbol{x}_{\mathrm{a}}\right)}\log \left[p\left(\boldsymbol{x}_{\mathrm{a}} \mid \boldsymbol{z}_{\mathrm{a}}\right)\right]}_{\circled{1}} \\
&+\underbrace{\mathbb{E}_{q\left(\boldsymbol{z}_{\mathrm{r}}, \boldsymbol{z}_{\mathrm{a}} \mid \boldsymbol{x}_{\mathrm{r}}, \boldsymbol{x}_{\mathrm{a}}\right)}\log \left[\frac{p\left(\boldsymbol{z}_{\mathrm{r}}, \boldsymbol{z}_{\mathrm{a}}\right)}{q(\boldsymbol{z}_{\mathrm{r}} \mid \boldsymbol{x}_{\mathrm{r}}) q\left(\boldsymbol{z}_{\mathrm{a}} \mid \boldsymbol{x}_{\mathrm{a}}\right)}\right]}_{\circled{2}}.
\end{aligned}
\label{eq:joint1}
\end{equation}
As we can see, the $\circled{1}$ part can be decomposed into two reconstruction losses, i.e., $\circled{1} = \mathcal{L}_{\mathrm{R}}^{\mathrm{r}} + \mathcal{L}_{\mathrm{R}}^{\mathrm{a}}$ and 
the $\circled{2}$ part in E.q. \ref{eq:joint1} can be decomposed as:

\begin{equation}
\scalebox{0.9}{
$\begin{aligned}
& \mathbb{E}_{q\left(\boldsymbol{z}_{\mathrm{r}}, \boldsymbol{z}_{\mathrm{a}} \mid \boldsymbol{x}_{\mathrm{r}}, \boldsymbol{x}_{\mathrm{a}}\right)} \log \left[\frac{p\left(\boldsymbol{z}_{\mathrm{r}}, \boldsymbol{z}_{\mathrm{a}}\right)}{q(\boldsymbol{z}_{\mathrm{r}} \mid \boldsymbol{x}_{\mathrm{r}}) q\left(\boldsymbol{z}_{\mathrm{a}} \mid \boldsymbol{x}_{\mathrm{a}}\right)}\right] \\
= & \mathbb{E}_{q\left(\boldsymbol{z}_{\mathrm{r}}, \boldsymbol{z}_{\mathrm{a}} \mid \boldsymbol{x}_{\mathrm{r}}, \boldsymbol{x}_{\mathrm{a}}\right)} \log \left[\frac{p\left(\boldsymbol{z}_{\mathrm{r}}, \boldsymbol{z}_{\mathrm{a}}\right) p(\boldsymbol{z}_{\mathrm{r}}) p\left(\boldsymbol{z}_{\mathrm{a}}\right)}{q(\boldsymbol{z}_{\mathrm{r}} \mid \boldsymbol{x}_{\mathrm{r}}) q\left(\boldsymbol{z}_{\mathrm{a}} \mid \boldsymbol{x}_{\mathrm{a}}\right) p(\boldsymbol{z}_{\mathrm{r}}) p\left(\boldsymbol{z}_{\mathrm{a}}\right)}\right] \\
 = & \mathbb{E}_{q\left(\boldsymbol{z}_{\mathrm{r}}, \boldsymbol{z}_{\mathrm{a}} \mid \boldsymbol{x}_{\mathrm{r}}, \boldsymbol{x}_{\mathrm{a}}\right)} \log \left[\frac{p\left(\boldsymbol{z}_{\mathrm{r}}, \boldsymbol{z}_{\mathrm{a}}\right)}{p(\boldsymbol{z}_{\mathrm{r}}) p\left(\boldsymbol{z}_{\mathrm{a}}\right)}\right] \\
 & + \mathbb{E}_{q\left(\boldsymbol{z}_{\mathrm{r}}, \boldsymbol{z}_{\mathrm{a}} \mid \boldsymbol{x}_{\mathrm{r}}, \boldsymbol{x}_{\mathrm{a}}\right)} \log \left[\frac{p(\boldsymbol{z}_{\mathrm{r}}) p\left(\boldsymbol{z}_{\mathrm{a}}\right)}{q(\boldsymbol{z}_{\mathrm{r}} \mid \boldsymbol{x}_{\mathrm{r}}) q\left(\boldsymbol{z}_{\mathrm{a}} \mid \boldsymbol{x}_{\mathrm{a}}\right)}\right] \\
 = & \underbrace{\mathbb{E}_{q\left(\boldsymbol{z}_{\mathrm{r}}, \boldsymbol{z}_{\mathrm{a}} \mid \boldsymbol{x}_{\mathrm{r}}, \boldsymbol{x}_{\mathrm{a}}\right)} \log \left[\frac{p\left(\boldsymbol{z}_{\mathrm{r}}, \boldsymbol{z}_{\mathrm{a}}\right)}{p(\boldsymbol{z}_{\mathrm{r}}) p\left(\boldsymbol{z}_{\mathrm{a}}\right)}\right]}_{\circled{A}} \\
 & - \underbrace{D_{KL}[q(\boldsymbol{z}_{\mathrm{r}} \mid \boldsymbol{x}_{\mathrm{r}}) \| p(\boldsymbol{z}_{\mathrm{r}})]}_{\circled{i}}-\underbrace{D_{KL}\left[q\left(\boldsymbol{z}_{\mathrm{a}} \mid \boldsymbol{x}_{\mathrm{a}}\right) \| p\left(\boldsymbol{z}_{\mathrm{a}}\right)\right]}_{\circled{ii}},
\end{aligned}$}
\label{eq:joint2}
\end{equation}
where the $\circled{2}$ part can be the combination of two inference losses and mutual information between latent variables, i.e., $\circled{2} = \mathcal{L}_{\mathrm{I}}^{\mathrm{r}} + \mathcal{L}_{\mathrm{I}}^{\mathrm{a}} + I(\boldsymbol{z}_{\mathrm{r}},\boldsymbol{z}_{\mathrm{a}})$,
where we denote $ \circled{i} = \mathcal{L}_{\mathrm{I}}^{\mathrm{r}}$, $ \circled{ii} = \mathcal{L}_{\mathrm{I}}^{\mathrm{a}}$, and $ \circled{A} = I(\boldsymbol{z}_{\mathrm{r}},\boldsymbol{z}_{\mathrm{a}})$.
\par
Minimization of the term denoted by $\circled{1}$ leads to an increased log-likelihood for both $p(\boldsymbol{x}_{\mathrm{r}}|\boldsymbol{z}_{\mathrm{r}})$ and $ p(\boldsymbol{x}_{\mathrm{a}}|\boldsymbol{z}_{\mathrm{a}})$, applicable to the raw and augmented data perspectives, respectively. Reducing the inference loss, represented as $\mathcal{L}_{\mathrm{I}}^{\mathrm{r}}$, $\mathcal{L}_{\mathrm{I}}^{\mathrm{a}}$ within the section labeled $\circled{2}$, contributes to a more coherent latent space that facilitates the reconstruction process. Moreover, enhancing the mutual information, denoted as $I(\boldsymbol{z}_{\mathrm{r}},\boldsymbol{z}_{\mathrm{a}})$, serves to bridge the disparities between the raw and augmented models. This process ensures a cohesive framework for incorporating data augmentation within the generative model. In conclusion, the proposed objective for learning is to approximate the joint distribution $p(\boldsymbol{x}_{\mathrm{r}},\boldsymbol{x}_{\mathrm{a}})$ within an augmentation-informed generative modeling context, denoted as:
\begin{equation}
    \begin{aligned}
        \mathcal{L}_{\mathrm{AVAE}} =& \circled{1} + \circled{A} - \circled{i}- \circled{ii} \\
        =& \mathcal{L}^{\mathrm{r}}_{\mathrm{ELBO}} + \mathcal{L}^{\mathrm{a}}_{\mathrm{ELBO}}  + I(\boldsymbol{z}_{\mathrm{r}},\boldsymbol{z}_{\mathrm{a}}),
    \end{aligned}
\label{eq:avae}
\end{equation}
where $\mathcal{L}_{\mathrm{AVAE}}$ represents the augmentation based VAE loss.
\begin{figure}[htbp]
    \centering
    \includegraphics[width=1\columnwidth]{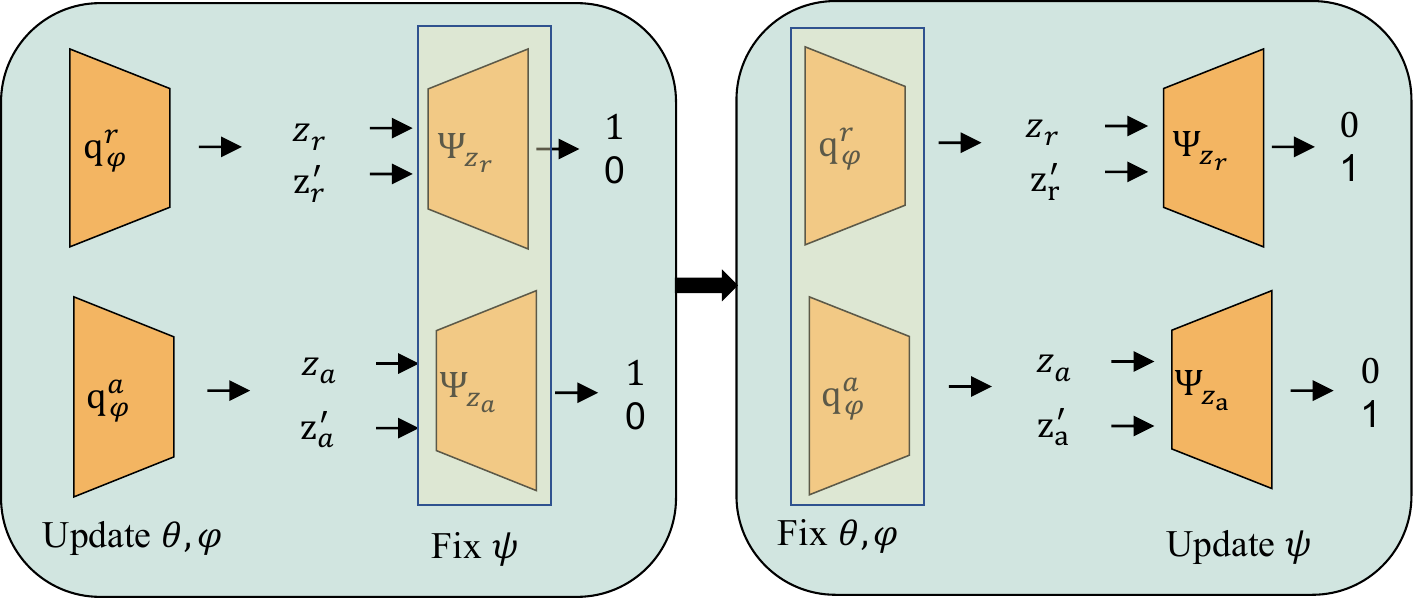}
    \caption{Illusration of adversarial learning in mutation information approximation. In the first stage, the discriminator is frozen to update the parameters of Encoders and decoders. In the second stage, We freeze the parameters of both the generator and the discriminator while simultaneously inverting the pseudo-labels of positive and negative samples to train the discriminator. }
    \label{fig:adversial}
\end{figure}
\subsection{Deep and Shallow Learning in Mutual Information Approximation}

\subsubsection{MI approximation in Shallow Learning}

We employ a $\mathcal{L}_{\mathrm{infoNCE}}$ loss to approximate the lower bound of MI. Since this method uses a non-parametric variational distribution in variational inference, it can be considered a form of shallow learning. When the variational distribution $q(\boldsymbol{z}_{\mathrm{r}}|\boldsymbol{z}_{\mathrm{a}})$ is employed to approximate the untractable posterior distribution $p(\boldsymbol{z}_{\mathrm{r}}|\boldsymbol{z}_{\mathrm{a}})$, as in E.q. (\ref{eq:infoNCE_appro-a}), we can derive a lower bound, as in E.q. (\ref{eq:infoNCE_appro-b}). Specifically, by using an energy-based variational function $q(\boldsymbol{z}_{\mathrm{r}}|\boldsymbol{z}_{\mathrm{a}})=\frac{p(\boldsymbol{z}_{\mathrm{r}})}{a(\boldsymbol{z}_{\mathrm{a}})} e^{f(\boldsymbol{z}_{\mathrm{r}}, \boldsymbol{z}_{\mathrm{a}})}$, where $f(\boldsymbol{z}_{\mathrm{r}}, \boldsymbol{z}_{\mathrm{a}})$ is a critic value function, and $a(\boldsymbol{z}_{\mathrm{a}})=\mathbb{E}_{p(x)}\left[e^{f(x, y)}\right]$ for approximation, we use the convexity of the $\log$ function to apply Jensen's inequality to $\mathbb{E}_{p(\boldsymbol{z}_{\mathrm{a}})}[\log a(\boldsymbol{z}_{\mathrm{a}})]$ to further derive a lower bound, as in E.q. (\ref{eq:infoNCE_appro-c}). By utilizing the inequality: $\log (\boldsymbol{z}) \leq \frac{\boldsymbol{z}}{\tau}+\log (\tau)-1$, we can approximate further to obtain another lower bound, as in E.q. (\ref{eq:infoNCE_appro-d}). Using $K$ samples for an unbiased estimate, we obtain E.q. (\ref{eq:infoNCE_appro-e}), and through Monte Carlo estimation, we can approximate it to the infoNCE loss, that is, $\mathcal{L}_{\mathrm{infoNCE}}$ in (\ref{eq:infoNCE_appro-f}):
\begin{subequations}
\label{eq:infoNCE_appro}
\begin{align}
& I(\boldsymbol{z}_{\mathrm{r}},\boldsymbol{z}_{\mathrm{a}})  \notag \\
= & \mathbb{E}_{p(\boldsymbol{z}_{\mathrm{r}}, \boldsymbol{z}_{\mathrm{a}})}\left[\log \frac{q(\boldsymbol{z}_{\mathrm{r}} \mid \boldsymbol{z}_{\mathrm{a}})p(\boldsymbol{z}_{\mathrm{r}}|\boldsymbol{z}_{\mathrm{a}})}{p(\boldsymbol{z}_{\mathrm{r}})q(\boldsymbol{z}_{\mathrm{r}} \mid \boldsymbol{z}_{\mathrm{a}})}\right] \label{eq:infoNCE_appro-a} \\
= & \mathbb{E}_{p(\boldsymbol{z}_{\mathrm{r}}, \boldsymbol{z}_{\mathrm{a}})}\left[\log \frac{q(\boldsymbol{z}_{\mathrm{r}} \mid \boldsymbol{z}_{\mathrm{a}})}{p(\boldsymbol{z}_{\mathrm{r}})}\right]  \notag \\ 
& +  \mathbb{E}_{p(\boldsymbol{z}_{\mathrm{a}})}[D_{\mathrm{KL}}(p(\boldsymbol{z}_{\mathrm{r}} \mid \boldsymbol{z}_{\mathrm{a}}) \| q(\boldsymbol{z}_{\mathrm{r}} \mid \boldsymbol{z}_{\mathrm{a}}))] \notag\\
\geq & \mathbb{E}_{p(\boldsymbol{z}_{\mathrm{r}}, \boldsymbol{z}_{\mathrm{a}})}[\log q(\boldsymbol{z}_{\mathrm{r}} \mid \boldsymbol{z}_{\mathrm{a}})] \label{eq:infoNCE_appro-b} \\
\geq & \mathbb{E}_{p(\boldsymbol{z}_{\mathrm{r}}, \boldsymbol{z}_{\mathrm{a}})}[f(\boldsymbol{z}_{\mathrm{r}}, \boldsymbol{z}_{\mathrm{a}})]  \notag\\
 & - \mathbb{E}_{p(\boldsymbol{z}_{\mathrm{a}})}\left[\frac{\mathbb{E}_{p(\boldsymbol{z}_{\mathrm{r}})}\left[e^{f(\boldsymbol{z}_{\mathrm{r}}, \boldsymbol{z}_{\mathrm{a}})}\right]}{a(\boldsymbol{z}_{\mathrm{a}})}+\log (a(\boldsymbol{z}_{\mathrm{a}}))-1\right] \label{eq:infoNCE_appro-c} \\
\geq  & 1  - \mathbb{E}_{p\left(\boldsymbol{z}_{(\mathrm{r}, 1: K)}\right) p(\boldsymbol{z}_{\mathrm{a}})}\left[\frac{e^{f\left(\boldsymbol{z}_{(\mathrm{r},1)}, \boldsymbol{z}_{\mathrm{a}}\right)}}{a\left(\boldsymbol{z}_{\mathrm{a}} ; \boldsymbol{z}_{(\mathrm{r}, 1: K)}\right)}\right] \notag\\
& +  \mathbb{E}_{p\left(\boldsymbol{z}_{(\mathrm{r}, 1: K)}\right) p\left(\boldsymbol{z}_{\mathrm{a}} \mid \boldsymbol{z}_{(\mathrm{r},1)}\right)}\left[\log \frac{e^{f\left(\boldsymbol{z}_{(\mathrm{r},1)}, \boldsymbol{z}_{\mathrm{a}}\right)}}{a\left(\boldsymbol{z}_{\mathrm{a}} , \boldsymbol{z}_{(\mathrm{r},1: K)}\right)}\right] \label{eq:infoNCE_appro-d}\\
\geq & \mathbb{E}\left[\frac{1}{K} \sum_{i=1}^K \log \frac{e^{f\left(\boldsymbol{z}_{(\mathrm{r}, i)}, \boldsymbol{z}_{(\mathrm{a}, i)}\right)}}{\frac{1}{K} \sum_{j=1}^K e^{f\left(\boldsymbol{z}_{(\mathrm{r}, i)}, \boldsymbol{z}_{(\mathrm{a}, j)}\right)}}\right]  \label{eq:infoNCE_appro-e}\\
\geq & \mathbb{E}\left[\frac{1}{K} \sum_{i=1}^K \log \frac{p\left(\boldsymbol{z}_{(\mathrm{a}, i)} \mid \boldsymbol{z}_{(\mathrm{r}, i)}\right)}{\frac{1}{K} \sum_{j=1}^K p\left(\boldsymbol{z}_{(\mathrm{a}, i)} \mid \boldsymbol{z}_{(\mathrm{r}, j)}\right)}\right] \triangleq \mathcal{L}_{\mathrm{infoNCE}}. \label{eq:infoNCE_appro-f}
\end{align}
\end{subequations}
Actually, we optimize an infoNCE loss scaled by the temperature coefficient $\tau$:
\begin{equation}
\begin{aligned}
&\mathcal{L}_{\text {InfoNCE }} \\
=& - \log \frac{\exp \left(\boldsymbol{z}_u^{\mathrm{r},\top} \boldsymbol{z}_u^{\mathrm{a}} / \tau\right)}{\sum_v \exp \left(\boldsymbol{z}_u^{\mathrm{r},\top} \boldsymbol{z}_v^{\mathrm{a}} / \tau\right)+\sum_{v \neq u} \exp \left(\boldsymbol{z}_u^{\mathrm{r},\top} \boldsymbol{z}_v / \tau\right)},
\end{aligned}
\end{equation}
where and the negative pairs are none, indicating that the negative keys for a sample are the positive keys for others.

\subsubsection{MI approximation in Deep Learning}
We can decompose the mutation information into two ratios in E.q. (\ref{eq:discri_appro-a}) and approximate in density ratio trick, guided by \cite{cheng2020club}. In that case, the density ratio is approached by a parameterized neural network, and we can approximate the MI implicitly in a deep learning scheme. Specifically, instead of modeling two distributions directly, i.e., $q(\boldsymbol{z}_{\mathrm{r}},\boldsymbol{z}_{\mathrm{a}})$ and $q(\boldsymbol{z}_{\mathrm{r}})$, we can learn the ratio $r = \frac{q(\boldsymbol{z}_{\mathrm{r}},\boldsymbol{z}_{\mathrm{a}})}{q(\boldsymbol{z}_{\mathrm{r}})}$ in an adversival manner, i.e., training a discriminator to classify whether the label comes from the target distribution $\mathcal{P}$ or not, as shown in E.q. (\ref{eq:discri_appro-b}), where the $y$ is a preset pseudo-label. Since we use the discriminator method to estimate the mutual information, the upper bound is denoted as E.q. (\ref{eq:discri_appro-c}).

\begin{subequations}
\label{eq:discri_appro}
\begin{align}
& \mathbb{E}_{q\left(\boldsymbol{z}_{\mathrm{r}}, \boldsymbol{z}_{\mathrm{a}}\right)} \frac{q\left(\boldsymbol{z}_{\mathrm{r}}| \boldsymbol{z}_{\mathrm{a}}\right)}{ q\left(\boldsymbol{z}_{\mathrm{a}}\right)} \notag \\
= & \mathbb{E}_{q\left(\boldsymbol{z}_{\mathrm{r}}, \boldsymbol{z}_{\mathrm{a}}\right)} \log \frac{q\left(\boldsymbol{z}_{\mathrm{a}}| \boldsymbol{z}_{\mathrm{a}}\right)}{q(\boldsymbol{z}_{\mathrm{r}})} \label{eq:discri_appro-a} \\
\leq &  \log \frac{\mathcal{P}(y=1 \mid \boldsymbol{z}_{\mathrm{r}})}{\mathcal{P}(y=0 \mid \boldsymbol{z}_{\mathrm{r}})}+\log \frac{\mathcal{P}(y=1 \mid \boldsymbol{z}_{\mathrm{a}})}{\mathcal{P}(y=0 \mid \boldsymbol{z}_{\mathrm{a}})} \label{eq:discri_appro-b} \\
\leq  & \log \frac{\Psi(\boldsymbol{z}_{\mathrm{r}})}{1-\Psi(\boldsymbol{z}_{\mathrm{r}})}+\log \frac{\Psi_{\mathrm{a}}(\boldsymbol{z}_{\mathrm{a}})}{1-\Psi_{\mathrm{a}}(\boldsymbol{z}_{\mathrm{a}})} \triangleq \mathcal{L}_{\mathrm{adversial}}\label{eq:discri_appro-c}
\end{align}
\end{subequations}

\subsection{End-to-End Anomaly Detection Training}

This section proposes an end-to-end TSAD model based on a weakly augmented generative model.
\subsubsection{Weakly Augmentation} In augmentation-based generative models, the likelihood fitting is enhanced by reusing training data. In VAEs, this leads to improved generative models $p_{\theta}(\boldsymbol{x}\mid\boldsymbol{z})$ parameterized by $\theta$ via enriched data representations in the inference network $q_{\phi}(\boldsymbol{z}\mid\boldsymbol{x})$ parameterized by $\phi$. The augmented latent variable, $\boldsymbol{z}_{\mathrm{a}}$, is derived as $\boldsymbol{z}_{\mathrm{a}} \sim q(\boldsymbol{z}_{\mathrm{a}}|\boldsymbol{x})$. During data preprocessing, we can augment latent representations directly by manipulating the input data augmentation, represented as $\boldsymbol{z}_{\mathrm{a}} \sim q(\boldsymbol{z}_{\mathrm{a}}|\boldsymbol{x}_{\mathrm{a}})$. Here, the augmented input $\boldsymbol{x}_{\mathrm{a}}$ is obtained from the raw input $\boldsymbol{x}_{\mathrm{r}}$ using the augmentation operation $\mathcal{O}$, formulated as $\boldsymbol{x}_{\mathrm{a}} = \mathcal{O}(\boldsymbol{x}_{\mathrm{r}})$.
\par
Data augmentation methods for time series data typically require an input array of size \texttt{(batch, time\_steps, channel)}, with manipulations possible in the batch domain (such as jittering with noise, scaling, and normalization) or in the time domain (including window slicing and warping). Additionally, augmentations can be applied in the frequency domain. These techniques enrich the original dataset through various methods, effectively enhancing data diversity. This diversification is crucial for models to comprehend better and predict time series patterns.

Nevertheless, our findings suggest that applying weak augmentation to the original input data may yield more favorable outcomes for likelihood fitting in anomaly detection tasks. Specifically, weak augmentation involves subtle modifications to the data, primarily through different normalization techniques. These include:
\begin{itemize}
\item \textbf{Standardization}:
    \begin{equation}
   \boldsymbol{x}_{\mathrm{a}} = \mathcal{O}^{\mathrm{stand}}(\boldsymbol{x}_{\mathrm{r}})= \frac{\boldsymbol{x}_{\mathrm{r}} - \mu}{\sigma},
   \label{eq:no1}
   \end{equation}
\end{itemize}  
where, $\mu$ is the mean and $\sigma$ the standard deviation of the data.
\begin{itemize}
\item \textbf{Min-Max Normalization}:
    \begin{equation}
   \boldsymbol{x}_{\mathrm{a}} = \mathcal{O}^{\mathrm{mm}}(\boldsymbol{x}_{\mathrm{r}}) = \frac{\boldsymbol{x}_{\mathrm{r}} - \mathrm{min}(\boldsymbol{x}_{\mathrm{r}})}{\mathrm{max}(\boldsymbol{x}_{\mathrm{r}}) - \mathrm{min}(\boldsymbol{x}_{\mathrm{r}})},
   \label{eq:no2}
\end{equation}
\end{itemize}   
this scales the data to a specified range, such as 0 to 1.

Such moderate adjustments typically preserve the fundamental characteristics and trends of the time series. They enable the model to discern the core attributes of the data more effectively, thereby enhancing prediction accuracy. Furthermore, weak augmentation maintains the data's authenticity, mitigating the risk of over-distorting the original data structure. This is vital for preserving both the reliability and interpretability of the model.

\subsubsection{Training}
For the raw input data $\boldsymbol{x}_{\mathrm{r}}$, we first obtain its augmented variable $\boldsymbol{x}_{\mathrm{a}}$ during the data preprocessing phase. Then, we train the inference networks for both the raw and augmented perspectives, namely $q_{\phi_{\mathrm{r}}}(\boldsymbol{z}_{\mathrm{r}} \mid \boldsymbol{x}_{\mathrm{r}})$ and $q_{\phi_{\mathrm{a}}}(\boldsymbol{z}_{\mathrm{a}} \mid \boldsymbol{x}_{\mathrm{a}})$, as well as the generative networks $p_{\theta_{\mathrm{r}}}(\boldsymbol{x}_{\mathrm{r}} \mid \boldsymbol{z}_{\mathrm{r}})$ and $p_{\theta_{\mathrm{a}}}(\boldsymbol{x}_{\mathrm{a}} \mid \boldsymbol{z}_{\mathrm{a}})$. At the same time, we optimize the inference and reconstruction losses based on Eq. (\ref{eq:raw_eblo}) and E.q. (\ref{eq:aug_eblo}).

During the training phase, we employ a strategy of sharing parameters for joint likelihood consolidation in raw one $p(\boldsymbol{x}_{\mathrm{r}},\boldsymbol{z}_{\mathrm{r}})$ and augmented one $p(\boldsymbol{x}_{\mathrm{a}},\boldsymbol{z}_{\mathrm{a}})$ to align the reconstruction effects. This means allowing both the inference and reconstruction networks to share the same structure and parameters during training. Such an approach reduces the number of model parameters and increases the generalization of the model, enabling it to learn normal patterns of different data for reconstructing normal data.

In the variational inference of the joint distribution's posterior distribution, we maximize the mutual information of the two latent variables to encourage the original generator and the augmented generator to tend towards producing similar data distributions. In our actual optimization objectives, i.e., the surrogated loss $\mathcal{L}_{\mathrm{WAVAE}}$, we implemented two methods to control the divergence of the two likelihood distributions: the $\mathcal{L}_{\mathrm{WAVAE}}^{\mathrm{infoNCE}}$ loss based on contrastive learning:
\begin{equation}
    \begin{aligned}
        \mathcal{L}_{\mathrm{WAVAE}}^{\mathrm{infoNCE}} =& \circled{1} + \circled{A} - \circled{i}- \circled{ii} \\
        =& \mathcal{L}^{\mathrm{r}}_{\mathrm{ELBO}} + \mathcal{L}^{\mathrm{a}}_{\mathrm{ELBO}}  + I(\boldsymbol{z}_{\mathrm{r}},\boldsymbol{z}_{\mathrm{a}}) \\
        =& \mathcal{L}^{\mathrm{r}}_{\mathrm{ELBO}} + \mathcal{L}^{\mathrm{a}}_{\mathrm{ELBO}}  + \mathcal{L}_{\mathrm{infoNCE}}(\boldsymbol{z}_{\mathrm{r}},\boldsymbol{z}_{\mathrm{a}}),
    \end{aligned}
\label{eq:wavae-info}
\end{equation}
and the $\mathcal{L}_{\mathrm{WAVAE}}^{\mathrm{adversial}}$ loss based on adversarial learning:
\begin{equation}
    \begin{aligned}
        \mathcal{L}_{\mathrm{WAVAE}}^{\mathrm{adversial}} =& \circled{1} + \circled{A} - \circled{i}- \circled{ii} \\
        =& \mathcal{L}^{\mathrm{r}}_{\mathrm{ELBO}} + \mathcal{L}^{\mathrm{a}}_{\mathrm{ELBO}}  + I(\boldsymbol{z}_{\mathrm{r}},\boldsymbol{z}_{\mathrm{a}}) \\
        =& \mathcal{L}^{\mathrm{r}}_{\mathrm{ELBO}} + \mathcal{L}^{\mathrm{a}}_{\mathrm{ELBO}}  + \mathcal{L}_{\mathrm{adversial}}(\boldsymbol{z}_{\mathrm{r}},\boldsymbol{z}_{\mathrm{a}}),
    \end{aligned}
\label{eq:wavae-adver}
\end{equation}
and the discriminator's performance is optimized in an adversarial manner, with the specific optimization process illustrated in Fig. \ref{fig:adversial}. The first part focuses on maximizing the encoder-decoder capabilities, and the second part involves swapping pseudo-labels to maximize discriminator loss.

\begin{figure*}
    \centering
    \includegraphics[width=2\columnwidth]{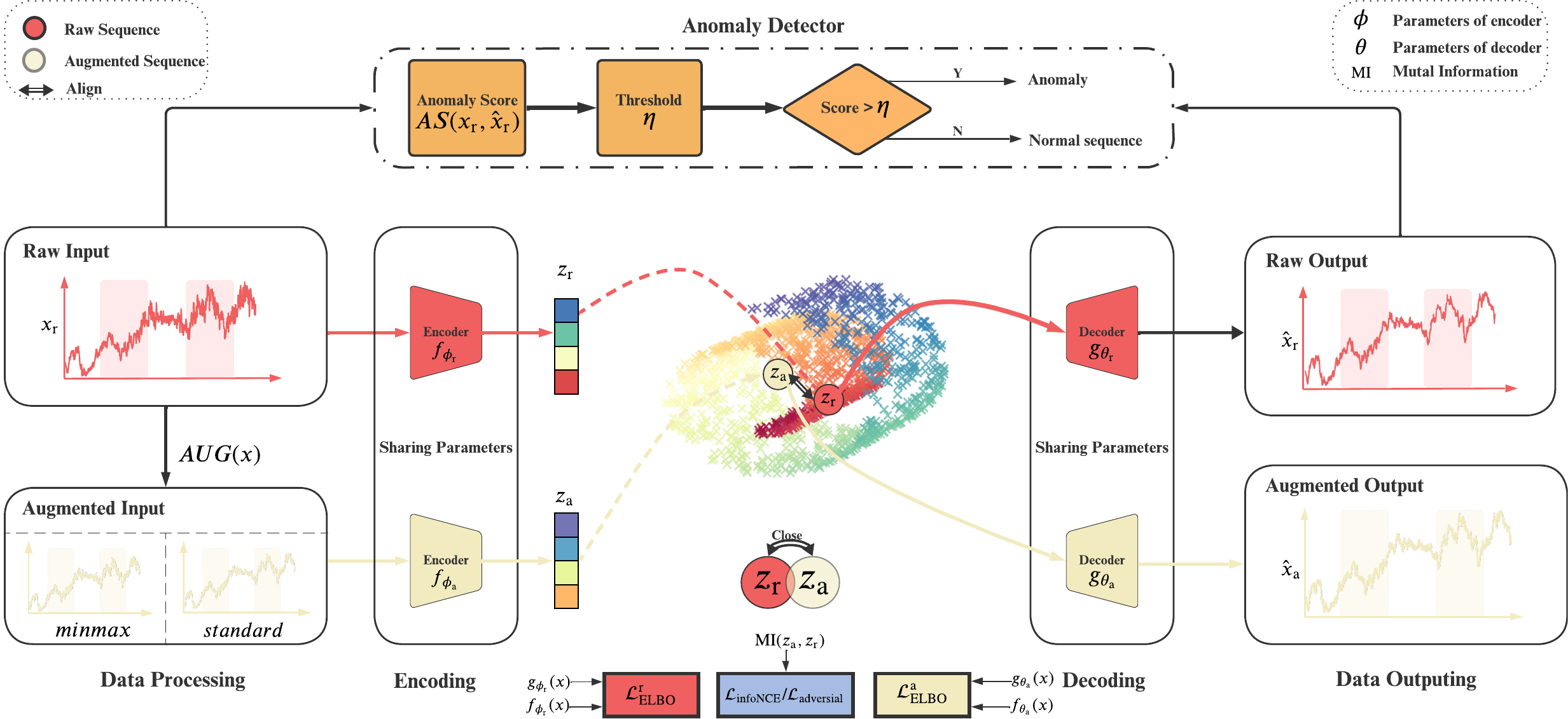}
    \caption{The overall framework of WAVAE, training begins with the raw data $\boldsymbol{x}_{\mathrm{r}}$ undergoing an augmentation algorithm $AUG$, resulting in augmented data $\boldsymbol{x}_{\mathrm{a}}$. Concurrently, we train a shared-parameter VAE separately for both sets of data. However, evaluation, i.e., Anomaly detector, is conducted solely on the original model between raw input 
 $\boldsymbol{x}_{\mathrm{r}}$ and its' reconstruction $\hat{\boldsymbol{x}}_{\mathrm{r}}$, essentially designing an end-to-end anomaly detector.}
    \label{fig:fra}
\end{figure*}

\subsubsection{Anomaly Scores} The training process and the determination of anomalies are illustrated in Fig. \ref{fig:fra}. Reconstruction-based anomaly detection utilizes the deviation between the original data and the reconstructed data as an anomaly score, denoted as $AS(\boldsymbol{x},\hat{\boldsymbol{x}})$. We determine whether the input data is anomalous by comparing the anomaly score with a pre-set threshold $\eta$. The specific process is shown in Algorithm \ref{algorithm:training}.

\begin{algorithm}[htbp]
\label{algorithm:training}
\caption{The training process of WAVAE}
\SetAlgoLined 
\KwIn{Dataset $\mathcal{D} = \{B_{tr}^{i}, B_{e}^{i}\}_{i=1}^{m}$, Training batch $B_{\mathrm{tr}} = \{(\boldsymbol{x}^{(j)}, \boldsymbol{y}^{(j})\}^{b}_{j=1}  \in \mathbb{R}^{b \times s \times f}$, Evaluation batch $B_{\mathrm{e}} \in \mathbb{R}^{b \times s \times c}$. \texttt{// $b, s, c$ is the size of batch, sequence length and features}}
\KwOut{Parameters of encoder $f_{\phi}$, decoder $g_{\theta}$, and discriminator $\psi$, Anomaly threshold $\eta$.}

\For{ each {$B^{i}$ in training batch $B_{tr}$}}{
    $B^{i}_{\mathrm{a}} = \mathcal{O}(B^{i})$; \texttt{// Here the operation $\mathcal{O}$ is defined in E.q. (\ref{eq:no1}) and E.q. (\ref{eq:no2}). } \\
    $B^{i}_{\mathrm{a}} \subset B_{\mathrm{a}}$;
}

\While{unconverged}{
    \For{each $B^{i},B^{i}_{\mathrm{a}}$ in $\mathcal{D}$}{
        Compute gradients of Eq. (\ref{eq:wavae-info}) or Eq. (\ref{eq:wavae-adver}) w.r.t. $\theta$ and $\phi$; \\
        Update the parameters of $f,g$;
    }
}

\For{ each $B^{i}_{e}$ in evaluation batch $B_{e}$}{
    $\hat{B^{j}_{e}} = g_{\theta}\big(f_{\phi}(B^{j}_{e})\big)$; \texttt{// Reconstruct the sequence} \\
    \For{each $\boldsymbol{x}_{\mathrm{r}}^{i}, \boldsymbol{x}_{\mathrm{a}}^{i}$ in $B_{e}$}{
        $Score = AS(\boldsymbol{x}_{\mathrm{r}}^{j}, \boldsymbol{x}_{\mathrm{a}}^{j})$; \texttt{//Calculate the anomaly score based on the similarity} \\
        \eIf{$Score < \eta$}{
            $\boldsymbol{x}_{\mathrm{r}}^{i}$  is an anomaly;
        }{
            $\boldsymbol{x}_{\mathrm{r}}^{i}$  is not an anomaly;
        }
    }
}
\end{algorithm}

\section{Experiments}
\label{sec:exp}

\subsection{Benchmarks}
To validate the effectiveness of our approach, we selected 16 reconstruction-based models for anomaly detection in time series data as benchmarks, which included 6 generative models (GANs and AEs based), which are specifically:
\begin{enumerate}
  \item Transformer Autoencoder (TAE) \cite{meng2020spacecraft}: A transformer autoencoder encodes and decodes time-series data to capture temporal dependencies.
  \item The Multi-scale CNN-RNN based Autoencoder (MSCREA): A CNN, RNN-based autoencoder leverages convolutional neural networks for representation extraction across multiple scales and recurrent neural networks for capturing temporal dependencies, tailored for enhancing anomaly detection in time-series data.
  \item BeatGAN (BGAN) \cite{zhou2019beatgan}: A GAN-based model for ECG anomaly detection, learning normal heartbeats to identify irregular patterns in time-series data.
  \item RNN-based Autoencoder (RAE) \cite{malhotra2016lstm}: A Gated Recurrent Unit (GRU) based autoencoder designed to encode and decode time-series data for anomaly detection by learning sequential patterns and temporal relationships.
  \item CNN-based Autoencoder \cite{zhang2019deep}: A CNN-based autoencoder architecture tailored for time-series analysis, utilizing convolutional layers to identify spatial patterns in data, essential for detecting anomalies in sequential datasets.
  \item RandNet (RN) \cite{chen2017outlier}: An ensemble of randomly structured autoencoders with adaptive sampling recognize efficiently and robustly detect outliers in data.
\end{enumerate}
and $10$ anomaly detection methods for time-series data based on probabilistic generative models, namely Variational Autoencoders, which are specifically:

\begin{enumerate}
  \item The Gaussian Mixture Model Variational Autoencoder (GMMVAE) \cite{liao2018unified}: VAE with GMM priors combines the probabilistic framework of Gaussian mixtures with the generative capabilities of VAEs to model complex distributions in time-series data, facilitating robust anomaly detection through learned latent representations.
  \item Variational Autoencoder (VAE) \cite{xu2018unsupervised}: A traditional VAE model to model the likelihood of generative data.
  \item Recurrent Neural Network based VAE (RNNVAE) \cite{park2018multimodal}: Merges Recurrent Neural Networks (RNN) with the variational approach to autoencoding, capturing temporal dependencies within sequential data for improved anomaly detection through stochastic latent spaces.
  \item The Variational RNN Autoencoder (VRAE) \cite{su2019robust}: The Variational RNN Autoencoder combines the sequence modeling strengths of RNNs with the probabilistic latent space of variational autoencoders, aiming to improve anomaly detection in time-series by learning complex temporal structures.
  \item $\alpha$-VQRAE, $\beta$-VQRAE, and $\gamma$-VQRAE \cite{kieu2022anomaly}: Extensions of VRAE, with RNN substituted by a Quasi-Recurrent Network and $\alpha$, $\beta$, $\gamma$-loglikelihood loss to help the VAE model achieve robust representation.
  \item $\alpha$-biVQRAE, $\beta$-biVQRAE, and $\gamma$-biVQRAE \cite{kieu2022anomaly}: Variants of VQRAE, with RNN extended to bilevel to achieve time dependence on time-series data, while the $\alpha$, $\beta$, $\gamma$-loglikelihood loss helps the VAE based model achieve robust representation.
\end{enumerate}


\subsection{Experiment Setup}

\begin{table*}[htbp]
\renewcommand{\arraystretch}{1.3}
\centering
\caption{Datasets overview.}
\label{tab:dataset}
\begin{tabular}{c||c|c|c|c|c|c|c|c}
\toprule[1pt]
\textbf{} & \textbf{Sequences} & \textbf{Anomaly Sequences} & \textbf{Avg. length} & \textbf{Avg. anomalies} & \textbf{Avg. Anomaly ratio (\%)} & \textbf{Features} & \textbf{Time Range} \\ \hline
\textbf{GD} & 5 & 1 & 16,220 & 39 & 0.24\% & 18 & in 760 ms \\ 
\textbf{HSS} & 4 & 2 & 19,634 & 4,517 & 23.01\% & 18 & in 14 ms \\ 
\textbf{ECG} & 1 & 1 & 500 & 31 & 6.2\% & 2 &  NA\\ 
\textbf{TD} & 1 & 1 & 1,000 & 362 & 36.2\% & 2 & NA \\ 
\textbf{S5\_A1} & 67 & 67 & 1,416 & 25 & 1.79\% & 1 & NA \\ 
\textbf{S5\_A2} & 100 & 100 & 1,421 & 5 & 0.35\% & 1 & NA \\ 
\textbf{S5\_A3} & 100 & 100 & 1,680 & 9 & 0.54\% & 1 & NA \\ 
\textbf{S5\_A4} & 100 & 100 & 1,680 & 8 & 0.48\% & 1 & NA \\ 
\toprule[1pt]
\end{tabular}
\end{table*}

\textbf{Datasets:}
To validate the effectiveness of our proposed methodology, we executed a series of experiments on a quartet of multivariate time series datasets: Genesis Demonstrator Data for Machine Learning, High Storage System Data for Energy Optimization, Electrocardiogram, and Trajectory Data, along with a single univariate time series dataset, the Yahoo S5. These datasets, encompassing several hundred temporal sequences, are sourced from real-world industrial systems or are synthetically generated, comprehensively evaluating the algorithm's performance.

The Genesis Demonstrator dataset for machine learning (GD)\footnote{\url{https://www.kaggle.com/datasets/inIT-OWL/genesis-demonstrator-data-for-machine-learning}} comprises 5 distinct sequences, encapsulating continuous or discrete signals recorded from portable pick-and-place robots at millisecond intervals. We harness the sequence replete with anomalies to target anomaly detection, specifically the \texttt{Genesis\_AnomalyLabels.csv}, which consists of 16,220 records. Within this framework, records marked with class 0 are designated as normal, whereas the remaining classifications, classes 1 and 2, are delineated as anomalies.

The High Storage System Data for Energy Optimization (HSS)\footnote{\url{https://www.kaggle.com/datasets/inIT-OWL/high-storage-system-data-for-energy-optimization}} dataset is composed of 4 sequences documenting the readings from induction sensors situated on conveyor belts. Anomaly detection is conducted on two labeled sequences: \texttt{HRSS\_anomalous\_standard.csv} and \texttt{HRSS\_anomalous\_optimized.csv}, together encompassing 23,645 records. Within each sequence, records tagged with class 0 are categorized as normal, whereas those labeled with class 1 are identified as anomalous.

The Electrocardiogram dataset (ECG)\footnote{\url{https://www.cs.ucr.edu/~eamonn/time_series_data_2018/}} is comprised of a solitary time-series sequence collected from PhysioNet signals attributed to a patient with severe congestive heart failure. For the sake of consistent comparison, we followed the guidelines proposed by researcher \cite{chen2015general}, utilizing \texttt{ECG5000\_TRAIN.tsv} from the training datasets for anomaly detection. This approach involves classifying three classes (Supraventricular ectopic beats, PVC, and Unclassifiable events) as anomalies, while the two remaining classes (R-on-T Premature Ventricular Contraction (PVC), Normal) are maintained as the normative data.

The Trajectory Data (TD)\footnote{\url{https://www.cs.ucr.edu/~eamonn/time_series_data_2018/}} dataset encapsulates a unique time-series sequence, with each data point being two-dimensional. These points represent the detection algorithm's accuracy in delineating the skeletal structure of a hand, coupled with assessments from three human evaluators on the algorithm's predictive accuracy. We undertake an unsupervised anomaly detection task in this setting using the \texttt{HandOutlines\_TRAIN.tsv} file extracted from the training set, comprising 1000 instances. Within this dataset, instances classified as normal bear the label of class 1, and those recognized as anomalies carry the label of class 0.

The Yahoo S5 (S5)\footnote{\url{https://webscope.sandbox.yahoo.com/catalog.php?datatype=s&did=70}} dataset encompasses an array of both authentic and synthetic time-series sequences. The synthetic component of this collection is distinguished by sequences that display diverse trends, degrees of noise, and seasonal patterns. Conversely, the real segment of the dataset encapsulates sequences that chronicle the performance metrics of assorted Yahoo services. This compilation contains a total of 367 labeled anomalies across both real and synthetic time series. The dataset is segmented into four distinct subsets: the A1, A2, A3, and A4 benchmarks. Within this schema, entries marked with class 0 are categorized as normal, whereas those annotated with class 1 are identified as anomalous.

The comparison of each dataset concerning data volume, dimensionality, temporal span of sequence acquisition, and proportion of anomalies is illustrated in Table \ref{tab:dataset}.

\textbf{Anomaly Scores:}
The presented code snippet delineates a section of an anomaly detection algorithm employing a Variational Autoencoder (VAE). The VAE is tasked with modeling the probability distribution of the input data. During the anomaly detection phase, the input data is reconstructed, and the reconstruction error is computed as the sum of squared differences between the original data $\boldsymbol{x}_{\mathrm{r}}$ and the reconstructed data $\hat{\boldsymbol{x}}$. Anomalies are flagged by setting a threshold at the 99th percentile of the error distribution, isolating the top $1\%$ of instances with the highest errors as anomalies.

Performance metrics such as F1 score, precision, and recall are calculated to gauge the accuracy of the test. These metrics are critical in determining the anomaly detection mechanism's true positive rate (precision) and sensitivity (recall). Furthermore, the algorithm utilizes the confusion matrix $ \mathrm{C} $, the F1 score, and Cohen's kappa score $ \kappa $ to evaluate its performance comprehensively.

The utilization of the Area Under the Receiver Operating Characteristic curve (AUROC) and the Precision-Recall curve (PRAUC) is noteworthy:

\begin{itemize}
    \item AUROC is derived by plotting the False Positive Rate (FPR) against the True Positive Rate (TPR) and calculating the Area Under the Curve (AUC). It is a robust measure of the classifier's discriminative power, particularly in class imbalance.
    \item PRAUC is obtained by plotting precision against the recall and computing the AUC, providing valuable insight into the performance of the positive class, which is often the minority in anomaly detection tasks.
\end{itemize}

The advantages of AUROC and PRAUC include their ability to provide a holistic measure of model performance insensitive to threshold selection and their effectiveness in conditions of class imbalance. The AUROC reflects the likelihood that the classifier will rank a randomly chosen positive instance more highly than a negative one. On the other hand, the PRAUC focuses on the model's performance in the positive class, making it particularly useful for datasets with significant class imbalances. The calculation of these metrics is as follows:

\begin{itemize}
    \item The function \texttt{roc\_curve} computes the FPR and TPR for various threshold values, while the function \texttt{auc} calculates the area under the ROC curve to determine the AUROC.
    \item The function \texttt{precision\_recall\_curve} is employed to compute the precision and recall at different thresholds, with the \texttt{auc} function again used to calculate the area under the precision-recall curve, yielding the PRAUC.
\end{itemize}

Using these metrics, the model's performance can be evaluated not solely based on accuracy but also on its robustness against false anomaly classifications (precision) and its capability to identify all true anomalies (recall).

\textbf{Implementation details:}
Our experimental setup was standardized to ensure a level playing field and control for potential performance biases introduced by Pytorch Versions. The versions selected for all implementations were Python 3.7.16, PyTorch 1.1.0, NumPy 1.19.2, CUDA toolkit 10.0.130, and cuDNN 7.6.5. This approach guaranteed that the proposed and comparative algorithms were evaluated under equivalent computational environments. Our hardware setup included NVIDIA Quadro RTX 6000 GPUs with driver version 525.105.17 and CUDA version 12.0. Additionally, we incorporated a randomness control module, employing seed values to govern the stochasticity across computational units such as GPU, Python, and PyTorch.

\subsection{Performance}
The anomaly detection performance on five public datasets can be found in Tables \ref{tab:prauc},\ref{tab:rocauc}. In terms of \texttt{PRAUC}  and \texttt{ROCAUC}  metrics, our method outperformed the baseline across all datasets, regardless of whether they are AE and GAN-based generative models or VAE-based ones. Note that our comparative data originates from \cite{kieu2022anomaly}. Additionally, we observed that methods based on adversarial mechanisms generally underperform compared to those using contrastive loss.

\begin{table}[htbp]
\renewcommand{\arraystretch}{1.3}
\centering
\caption{Overall accuracy, PR-AUC. For each dataset, the three best-performing methods are denoted using distinct markings: \textbf{bold} for the top method, superscript asterisk$^{*}$ for the second-best, and \underline{underline} for the third-best. }
\label{tab:prauc}
\scalebox{0.98}{
\begin{tabular}{c||p{0.7cm}p{0.7cm}p{0.7cm}p{0.7cm}p{0.7cm}}
\toprule[1pt]
\textbf{Models/Datasets} & \textbf{GD} & \textbf{HSS} & \textbf{ECG} & \textbf{TD} & \textbf{S5} \\ \hline

 \textbf{TAE} & 0.088& 0.195 & 0.138& 0.175& 0.298\\
 \textbf{MSCREA} & 0.075& 0.161& 0.105&0.148 & N/A  \\
 \textbf{BGAN} & 0.109&0.214&0.103 &0.151 &0.434 \\
 \textbf{RAE} &0.128 &0.242 &0.118 &0.163 &0.421 \\
 \textbf{CAE} & 0.116&0.207 &0.107 &0.177 & 0.383\\

 \textbf{RN} &0.112 &0.146 &0.105 &0.168 &0.232 \\  \cline{1-6}
 \textbf{GMMVAE} & 0.142& 0.216& 0.163& 0.364& 0.458\\
 \textbf{VAE} & 0.097 &0.203 & 0.131 & 0.188 & 0.272 \\
 \textbf{RNNVAE} &0.086 & 0.204& 0.079& 0.118&0.211 \\
 \textbf{VRAE} & 0.131&0.219 & 0.144& 0.165& 0.298\\
 \textbf{$\alpha$-VQRAE} &0.235 & \underline{0.225} & 0.177&0.428 & 0.487\\
 \textbf{$\beta$-VQRAE} & 0.242& 0.223&0.177 &0.427& \underline{0.525}\\
 \textbf{$\gamma$-VQRAE} & 0.245& 0.222&0.184 & 0.423& 0.499\\
 \textbf{$\alpha$-biVQRAE} & 0.249&0.227$^{*}$& 0.141&0.429 & 0.490\\
 \textbf{$\beta$-biVQRAE} &\underline{0.256} & 0.224& 0.189$^{*}$& \underline{0.430}& 0.527$^{*}$\\
 \textbf{$\gamma$-biVQRAE} & 0.258$^{*}$& 0.222& \underline{0.186}& 0.432&  0.524\\\cline{1-6}
 \textbf{WAVQRAE-Adverisal} &\textbf{0.304} & \textbf{0.286}& \textbf{0.190} &\textbf{0.440} & \textbf{0.805}\\
 \textbf{WAVQRAE-Contrast} &\textbf{0.307} & \textbf{0.358}& \textbf{0.200} &\textbf{0.504} & \textbf{0.838}\\
 
\toprule[1pt]
\end{tabular}
}
\end{table}

\begin{table}[htbp]
\renewcommand{\arraystretch}{1.3}
\centering
\caption{Overall accuracy, ROC-AUC.}
\label{tab:rocauc}
\scalebox{0.98}{
\begin{tabular}{c||p{0.7cm}p{0.7cm}p{0.7cm}p{0.7cm}p{0.7cm}}
\toprule[1pt]
\textbf{Models/Datasets} & \textbf{GD} & \textbf{HSS} & \textbf{ECG} & \textbf{TD} & \textbf{S5} \\ \hline

 \textbf{TAE} & 0.652& 0.563$^{*}$ & 0.542& 0.531& 0.635\\
 \textbf{MSCREA} & 0.582& 0.509&0.509 & 0.519& N/A\\
 \textbf{BGAN} & 0.673&0.549 &0.547 & \underline{0.622}&0.677 \\
 \textbf{RAE} & 0.608&0.537 & 0.552& 0.593& 0.753\\
 \textbf{CAE} & 0.641& \underline{0.560}& 0.574& 0.583&0.757$^{*}$\\
 \textbf{RN} &0.731 &0.526 &0.524 &0.533 &0.575  \\ \cline{1-6}

 \textbf{GMMVAE} & 0.763&0.534 &0.533 & 0.531& 0.815\\
\textbf{VAE} & 0.664& 0.525& 0.531& 0.643& 0.678\\
 \textbf{RNNVAE} & 0.595& 0.516&0.536 &0.574 &0.642 \\
 \textbf{VRAE} & 0.658& 0.521&0.551 & 0.662$^{*}$& 0.660\\
 \textbf{$\alpha$-VQRAE} & 0.970& 0.529& 0.592& 0.539& 0.858\\
 \textbf{$\beta$-VQRAE} & 0.968& 0.520& 0.583& 0.535& 0.849\\
 \textbf{$\gamma$-VQRAE} & 0.969& 0.524&0.598 & 0.547& \underline{0.875}\\
 \textbf{$\alpha$-biVQRAE} & 0.975& 0.538& 0.597&0.542 & 0.873\\
 \textbf{$\beta$-biVQRAE} &\underline{0.976} &0.527 &0.603$^{*}$& 0.546& 0.864\\
 \textbf{$\gamma$-biVQRAE} & 0.978$^{*}$& 0.526& \underline{0.601}& 0.549&0.882$^{*}$ \\ \cline{1-6}
 \textbf{WAVQRAE-Adverisal} &\textbf{0.991} & \textbf{0.563} &\textbf{0.612} &\textbf{0.579} & \textbf{0.883}\\
 \textbf{WAVQRAE-Contrast} &\textbf{0.996} & \textbf{0.575} &\textbf{0.630} &\textbf{0.646} & \textbf{0.899}\\
\toprule[1pt]
\end{tabular}
}
\end{table}


\subsection{Sensitivity Analysis}

To rigorously assess the sensitivity of hyperparameters in our model, we have conducted an extensive series of ablation experiments. This comprehensive evaluation encompasses many hyperparameter sets throughout the entire end-to-end training process. 
\begin{itemize}
    \item Specifically, we investigate variations in VAE-related hyperparameters such as the $\beta$ in E.q. (\ref{eq:raw_eblo}) and E.q. (\ref{eq:aug_eblo}) to balance the inference and reconstruction in VAE training, the dimension of $\boldsymbol{z}$, and the reconstruction loss $\mathcal{L}_{\mathrm{R}}$.
    \item We also scrutinize SSL-related hyperparameters like the number of discriminator layers, the weight of infoVAE loss, and the augmentation method.
    \item In addition, we delve into hyperparameters pertinent to time-series processing, including the sequence length and hidden variables in embeddings.
    \item Lastly, we explore adjustments in deep learning hyperparameters, including batch size, learning rates, and the number of epochs. 
\end{itemize}   
In each set of experiments, we systematically vary a selected hyperparameter within its feasible range while maintaining the default settings for all other hyperparameters to isolate and understand the individual impact of each hyperparameter adjustment on the model's performance.

\begin{figure*}
    \centering
    \includegraphics[width=2\columnwidth]{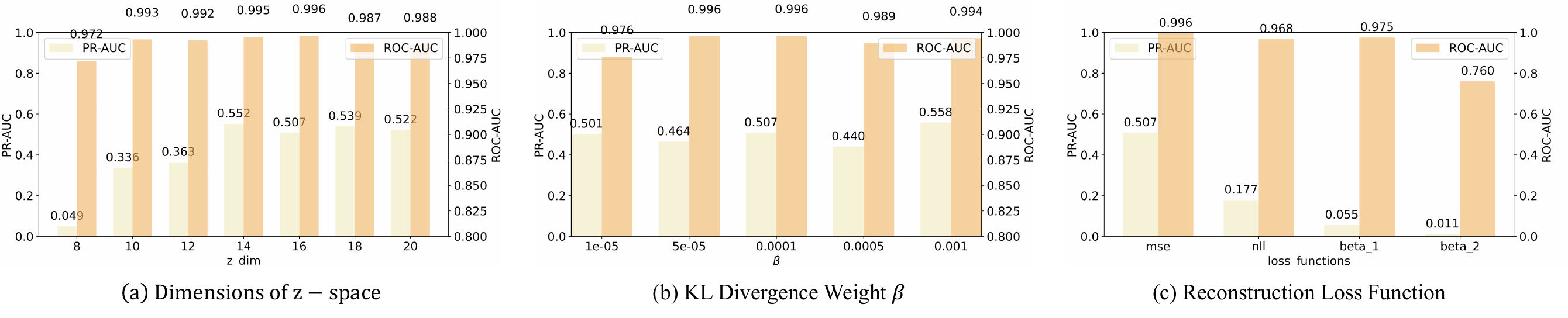}
    \caption{Sensitivity analysis of VAE related hyperparameters indicates significant findings: (a) reveals that the dimension of $\boldsymbol{z}$ profoundly influences outcomes, with optimal performance when the dimension ranges between 14 and 20. (b) shows that $\beta$ exerts a minimal effect on optimization, peaking in efficacy at 0.001. (c) demonstrates the superior performance of the MSE loss function.}
    \label{fig:abli_vae}
\end{figure*}

\begin{figure*}
    \centering
    \includegraphics[width=2\columnwidth]{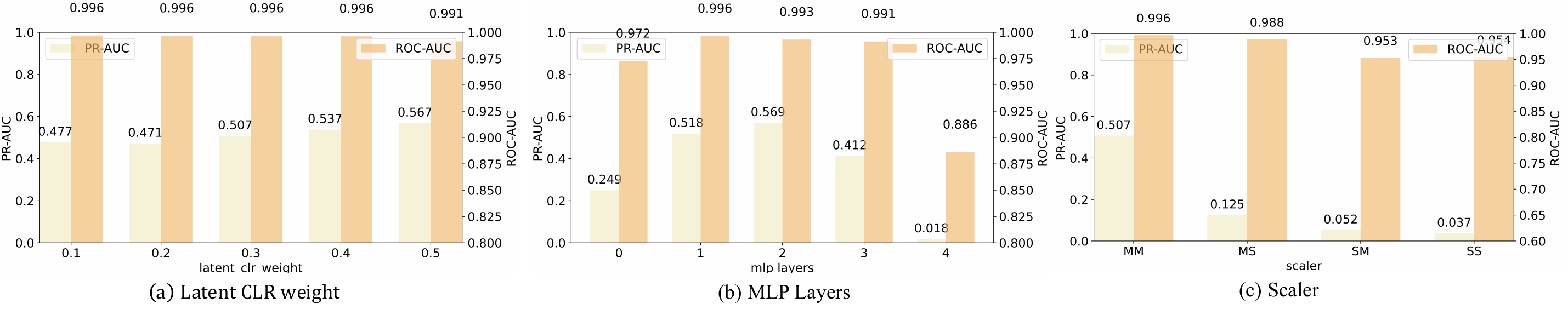}
    \caption{Sensitivity analysis of SSL loss related hyperparameters. From (a), it is observed that the weight of the infoNCE Loss has a minimal impact on the overall effectiveness. Conversely, (b) indicates that the number of layers in the discriminator significantly affects the results, with the best performance observed between 2 and 3 layers. (c) illustrates varying augmentation approaches, indicating that using min-max normalization (\texttt{MinMax}) on both original and augmented data is the most effective.}
    \label{fig:abli_ssl}
\end{figure*}

\begin{figure*}
    \centering
    \includegraphics[width=2\columnwidth]{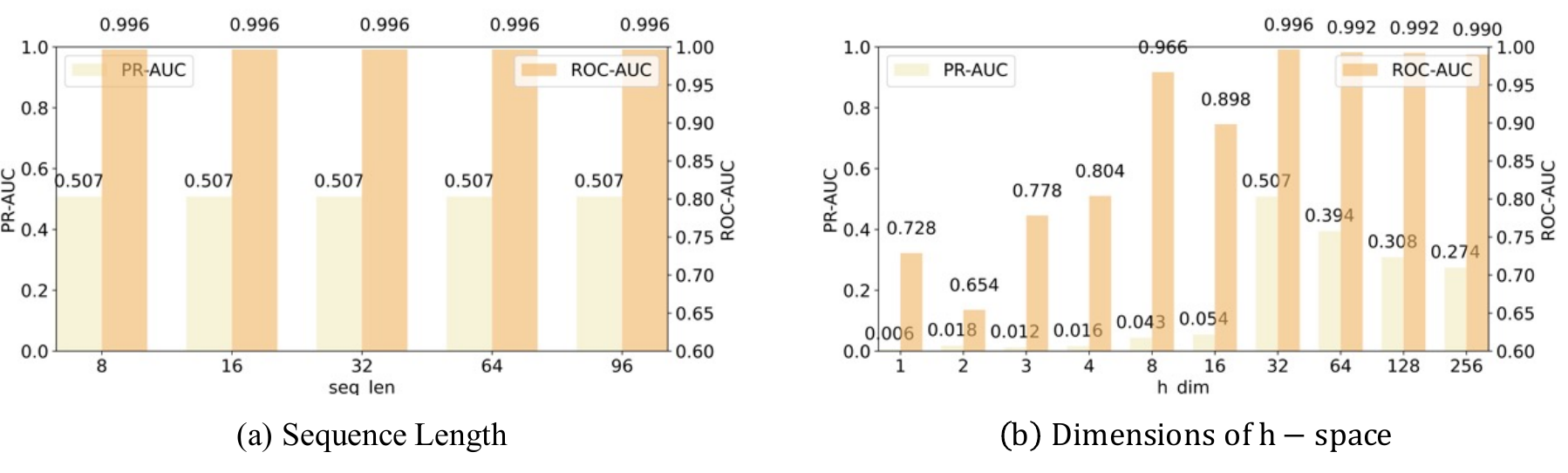}
    \caption{Sensitivity analysis of sequence-related hyperparameters. (a) indicates that the model's anomaly detection performance is not affected by the length of the series. (b) shows that the encoding network achieves the best performance when the hidden state size is 32. }
    \label{fig:abli_se}
\end{figure*}

\begin{figure*}
    \centering
    \includegraphics[width=2\columnwidth]{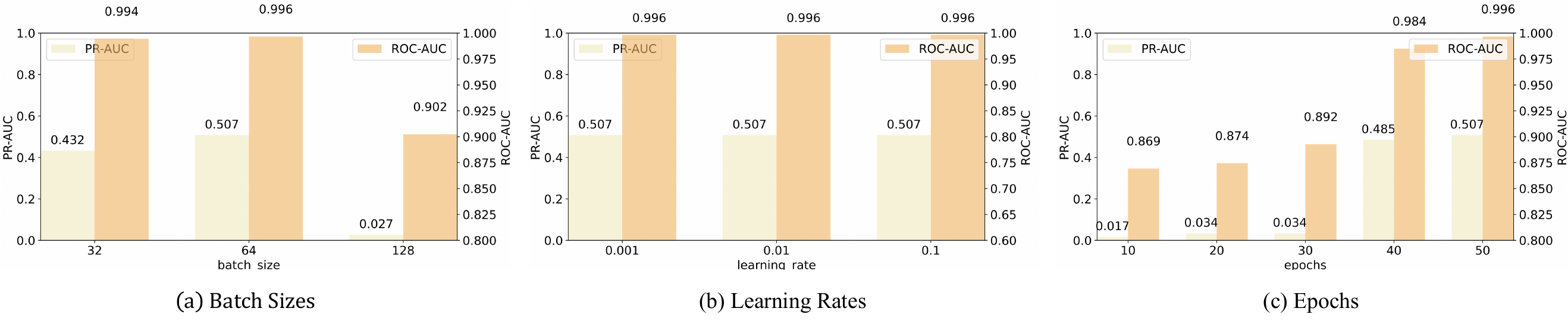}
    \caption{Sensitivity analysis of deep learning related hyperparameters. (a) A batch size of 64 yields optimal results. (b) The learning rate has minimal impact on the model. (c) The best performance was observed at 50 epochs.}
    \label{fig:abli_deep}
\end{figure*}

\subsubsection{Effect of VAEs}
Ideally, VAE is adept at modeling data distributions, encapsulating the potential to fit the likelihood of diverse data modalities through its sophisticated encoder-decoder architecture rooted in deep neural networks. Concurrently, it postulates a manifold-based, low-dimensional, continuous and smooth space. However, in real-world applications, the efficacy of a VAE's data likelihood estimation is subject to substantial variability, influenced by the selection of encoder-decoder architectures, the diversity of data modalities, and the specificities of the task at hand. To isolate and assess the effects of these factors on anomaly detection performance, we embark on a systematic sensitivity analysis of hyperparameters spanning three pivotal domains: weight of KL Divergence $\beta$, dimensions of latent variables $\boldsymbol{d}$, and reconstruction loss function. Through this methodical examination, we aim to elucidate the impact of these variables on the VAE's reconstruction proficiency, thereby enhancing the model's suitability for anomaly detection endeavors.
\par
\textbf{Dimensions of Latent variables:} 
The dimension of the latent variable determines the amount of information the encoder compresses to maximize the log-likelihood under the Information Bottleneck Theory. Simultaneously, it influences the dependency and causality of low-dimensional space representations under the manifold assumption. The fundamental assumptions of generative model-based time-series anomaly detection posit that anomaly data will deviate from the likelihood of normal data. We conducted experiments with varying dimensions of the latent variable $\boldsymbol{z}$ to develop a robust likelihood function, specifically exploring $\{8,10,12,14,16,18,20\}$. The outcomes and in-depth analysis of these experiments are detailed in Fig. \ref{fig:abli_vae} (a), demonstrating the optimal dimension for $\boldsymbol{z}$.
\par
\textbf{KL Divergence Weight:}
The KL weight controls the balance between representation learning and reconstruction in the VAE model and the information during the compression process, affecting the model's robustness during training. We use the hyperparameter $\beta$ to adjust the VAE's compression capability. We selected five distinct values for the KL term to assess their impact, specifically $\{1e-5, 5e-5, 1e-4, 5e-4, 1e-3\}$. Detailed results and analysis of this exploration are presented in Fig. \ref{fig:abli_vae} (b).
\par
\textbf{Reconstruction Loss Function:}
In the Eq. (\ref{eq:raw_eblo}) and E.q. (\ref{eq:aug_eblo}), we fit different likelihood distributions by optimizing the specific reconstruction loss. For discrete data, we optimize the Binary Cross Entropy (BCE) loss, i.e., $\mathcal{L}_{\mathrm{R}}^{\mathrm{BCE}}$ to fit the log-likelihood of a multivariate Bernoulli distribution, denoted as:
\begin{equation}
\begin{aligned}
    & \mathcal{L}_{\mathrm{R}}^{\mathrm{BCE}}  \\
    =&  E_{q_\phi\left(\boldsymbol{z} \mid \boldsymbol{x}\right)}\left[\log p_\theta\left(\boldsymbol{x} \mid \boldsymbol{z}\right)\right]\\
    =& E_{q_\phi\left(\boldsymbol{z} \mid \boldsymbol{x}\right)}\left[\sum_{d=1}^D \boldsymbol{x}_{d} \log \lambda_{\theta, d}\left(\boldsymbol{z}\right)+\left(1-\boldsymbol{x}_{d}\right) \log \left(1-\lambda_{\theta, d}\left(\boldsymbol{z}\right)\right)\right],
\end{aligned}
\end{equation}
Where $\boldsymbol{x} \in \{0,1\}^{D}$ and $\lambda \in \{0,1\}^{D}$ are the parameters of univariate Bernoulli distributions.
For continuous data, we optimize the Mean Square Error (MSE) $\mathcal{L}_{\mathrm{R}}^{\mathrm{MSE}}$ to fit the log-likelihood of a multivariate Gaussian distribution, denoted:
\begin{equation}
\begin{aligned}
     \mathcal{L}_{\mathrm{R}}^{\mathrm{MSE}}  
    =&  E_{q\left(\boldsymbol{z} \mid \boldsymbol{x}\right)}\left[\log p\left(\boldsymbol{x} \mid \boldsymbol{z}\right)\right]\\
    =&  \frac{1}{D} \sum_{d=1}^D||\boldsymbol{x}_{d} - \hat{\boldsymbol{x}}_{d} ||^2.
\end{aligned}
\end{equation}
We also tested two robust variants \cite{futami2018variational} based on the Bernoulli likelihood distribution:
\begin{equation}
\begin{aligned}
& \mathcal{L}_{\mathrm{R}}^{\mathrm{robust1}}  \\
    =&\frac{\alpha_{1}+1}{\alpha_{1}}\left(\prod_{d=1}^D\left(\boldsymbol{x}_d \hat{\boldsymbol{x}}_{d}^{\alpha_{1}}+\left(1-\boldsymbol{x}_d\right)\left(1-\hat{\boldsymbol{x}}_{d}\right)^{\alpha_{1}}\right)-1\right),
\end{aligned}
\end{equation}
and Gaussian likelihood distribution:
\begin{equation}
\begin{aligned}
    & \mathcal{L}_{\mathrm{R}}^{\mathrm{robust2}}  \\
    =& \frac{\alpha_{2}+1}{\alpha_{2}}\left(\frac{1}{\left(2 \pi \sigma^2\right)^{\alpha_{2} D / 2}} \exp \left(-\frac{\alpha_{2}}{2 \sigma^2} \sum_{d=1}^D\left\|\hat{\boldsymbol{x}}_d - \boldsymbol{x}_d\right\|^2\right)-1\right),
\end{aligned}
\end{equation}
where $\alpha_1,\alpha_1$ are the hyperparameters and $\sigma$ is the variance.
The analysis and comparison of four types of loss functions are illustrated in Fig. \ref{fig:abli_vae} (c).
\par
\subsubsection{Effect of SSL Loss}
The SSL Loss in E.q. (\ref{eq:wavae-info}) and E.q. (\ref{eq:wavae-adver}) will be biased by the approximation methods and augmentation types. 
\par
\textbf{Mutual Information Approximation:}
Our study investigated two loss functions for mutual information maximization: the infoNCE Loss in contrastive learning, with weight hyperparameters $\{0.1, 0.2, 0.3, 0.4, 0.5\}$, detailed in Fig. \ref{fig:abli_ssl} (a), and the adversarial learning discriminator, varying layers [3, 4, 5, 6], analyzed in Fig. \ref{fig:abli_ssl} (b).
\par
\textbf{Augumentation Methods:}
For self-supervised methods applied to time-series data, augmentation can be employed to mine the intrinsic characteristics of the data, addressing the issue of insufficient data for deep models. To validate the effectiveness of our approach, we experimented with various strong augmentations that enhance the time dependencies and frequency domain representations of time-series data. In parallel, we also explored several weak augmentations, specifically normalization techniques applied to time-series data. Our findings indicate that the domains transformed by strong augmentations are ill-suited for generating robust likelihood functions, leading to suboptimal results in anomaly detection. In that case, We conducted sensitivity analysis experiments by testing two combinations of weak augmentations. These combinations include both raw and augmented data using \texttt{MinMax} (Fig. \ref{fig:abli_ssl} (c) \texttt{MM}), raw data with \texttt{MinMax} and augmented data with \texttt{Standard} (Fig. \ref{fig:abli_ssl} (c) \texttt{MS}), raw data with \texttt{Standard} and augmented data with \texttt{MinMax} (Fig. \ref{fig:abli_ssl} (c) \texttt{SM}), and both raw and augmented data using standardization (Fig. \ref{fig:abli_ssl} (c) \texttt{SS}). Specific experimental results and analysis are presented in Fig. \ref{fig:abli_ssl} (c).
\subsubsection{Effect of Time Series Processing}
The time series' inherent characteristics, such as the window size in a batch and the memory step length in the encoding model, can impact model performance.
\par
\textbf{Sequence Length}
In time-series data analysis, the window length is critical as it sets the data truncation extent, which is essential for detecting anomalies with periodicity or spatio-temporal continuity. Furthermore, the length of the time series plays a significant role in identifying contextual anomalies. We chose time series lengths of $\{8, 16, 32, 64, 96\}$ for our sensitivity analysis. Detailed experimental results and analyses are illustrated in Fig. \ref{fig:abli_se} (a).
\par
\textbf{Hidden Vairbales:}
We evaluated the impact of different hidden space sizes in the embedding, experimenting with dimensions of $\{1, 2, 3, 4, 8, 16, 32, 64, 128,256\}$. Detailed analysis and results are presented in Fig. \ref{fig:abli_se} (b).
\subsubsection{Effect of Deep Learning}
In deep models, batch size, learning rates, and epochs cooperate to guide the model convergence to the optimal. 
\par
\textbf{Batch Sizes:}
By modulating the batch size, we gain insights into the stability of gradient updates and their consequent impact on model convergence. To this end, we selected batch sizes $\{32, 64, 128\}$ to empirically ascertain their influence on the model's performance. Detailed experimental results and analyses are illustrated in Fig. \ref{fig:abli_deep} (a).
\par
\textbf{Leanring Rates:}
Step size in gradient descent induced by the learning rates is taken during optimization and can significantly influence the model's ability to find minima. We test learning rates of $\{0.001, 0.01, 0.1\}$ and systematically study their effects and optimize the model's performance. Fig. \ref{fig:abli_deep} (b) illustrates detailed experimental results and analyses.
\par
\textbf{Number of Epochs:}
In the context of unsupervised anomaly detection, rather than focusing on model generalization, we prioritize the impact of training duration on performance. We fix the randomness and maintain consistent hyperparameters, testing the same model's anomaly detection capabilities at epochs $\{10, 20, 30, 40, 50\}$. Fig. \ref{fig:abli_deep} (c) illustrates detailed experimental results and analyses.

\section{Conclusion}
\label{sec:con}

The VAE-based anomaly detection effectively captures underlying data distributions in time series analysis. As a result, anomalies outside this distribution show notable reconstruction errors. However, the limited amount of data samples can affect the model's ability to fit this distribution, 
especially when considering the rare and hard-to-detect nature of anomalies in real-world situations. To combat data scarcity, we introduce a weakly augmented VAE for time series anomaly detection. The model can achieve a more robust representation in the latent space through joint training on augmented data. Meanwhile, we present two self-supervised strategies, adversarial and contrastive learning, to enhance the performance in data fitting. Quantitative experimental results demonstrate that our approach exhibits commendable performance across five datasets under two distinct foundational model architectures.

\newpage
\bibliography{vae}
\bibliographystyle{IEEEtran}

\end{document}